\documentclass[runningheads]{llncs}

\usepackage{eccv}

\usepackage{eccvabbrv}

\usepackage{graphicx}
\usepackage{booktabs}

\usepackage[accsupp]{axessibility}

\usepackage[pagebackref,breaklinks,colorlinks,citecolor=eccvblue]{hyperref}
\usepackage{hyperref}

\usepackage{orcidlink}

\usepackage{enumitem}  

\newcommand{\modelname}[1]{DR-GS}
\usepackage{booktabs}
\usepackage{bm} 

\usepackage{amsmath,amssymb}
\usepackage{algorithm}
\usepackage{algpseudocode}
\usepackage{caption}
\usepackage{float} 
\usepackage{colortbl} 
\usepackage{graphicx}
\usepackage{adjustbox}

\begin{document}

\title{DR-GS: Physically-Based Deformable and Relightable 2D Gaussians} 

\author{Jiaxin Li\inst{1,2,3} \and Tong Wu
\inst{5} \and Yi Wei \inst{5} \and Tailin Wu \inst{3} \and Li Zhang \inst{1,4}\thanks{Corresponding author. E-mail: \email{lizhangfd@fudan.edu.cn}} }

\authorrunning{J.~Li et al.}

\institute{Shanghai Innovation Institute, China \and
Zhejiang University, China \and
Department of Artificial Intelligence, Westlake University, China
 \and
School of Data Science, Fudan University, China
\and
Central Media Technology Institute, Huawei, China}

\maketitle

\begin{figure}[H]
    \centering
    \includegraphics[width=1\linewidth]{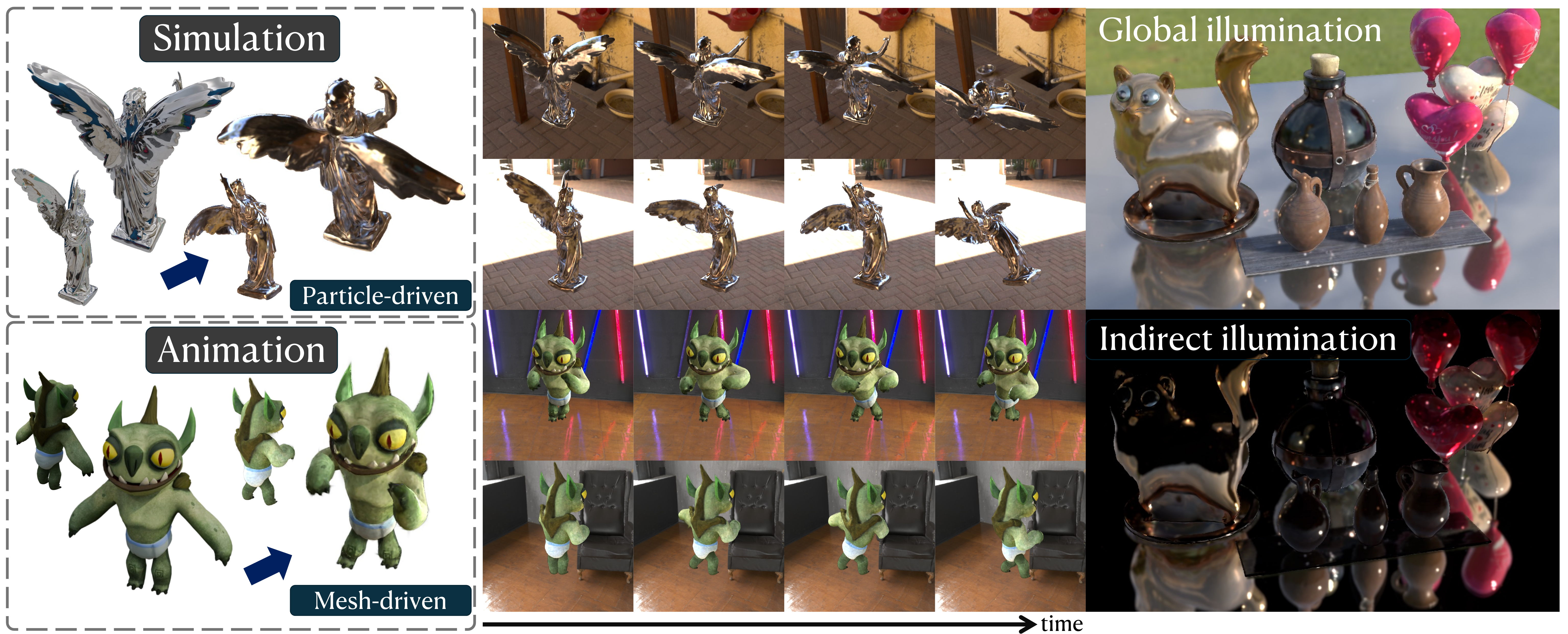} 
    \caption{
    \textbf{DR-GS} enables physically plausible rendering under deformations and lighting changes via decoupled geometry, lighting, and materials. The project page is available at \url{https://jiaxinlia.github.io/DR-GS/}.
    }
    \label{fig:teaser}
\end{figure}

\begin{abstract}

Gaussian splatting (GS) has garnered significant attention in VR/AR and digital content creation due to its explicit parameterization and efficient rendering capabilities. However, existing GS-based methods for deformable objects face two key limitations: \textbf{(i)} illumination is erroneously baked into textures, causing physically inconsistent responses under dynamic deformations and lighting changes; \textbf{(ii)} snapshot-based reconstruction restricts post-reconstruction material editing. To address these challenges, we propose \textbf{\underline{D}}eformable and \textbf{\underline{R}}elightable \textbf{\underline{GS}} (\textbf{\modelname{}}), a unified Gaussian framework that integrates physically-based inverse rendering, relighting, and deformation-aware manipulation. Through explicitly disentangling geometry, illumination, and material representations, \modelname{} overcomes the limitations of static snapshots, resolving unrealistic appearance under varying conditions while enabling post-reconstruction parameter editing. Extensive experiments show that \modelname{} achieves leading visual quality across static reconstruction, dynamic deformation, and relighting, reliably preserving reflections and specular highlights on glossy surfaces. It further establishes a fully decoupled geometry-illumination-material pipeline, enabling high-quality 3D asset creation and comprehensive post-editing. 

\end{abstract}
\section{Introduction}
Gaussian splatting (GS) represents 3D scenes with learnable anisotropic Gaussians, providing an explicit scene representation and real-time rendering. Compared with NeRF~\cite{2020nerf}, GS achieves substantially higher rendering throughput and supports interactive applications that require both realism and low latency, such as VR/AR~\cite{vr-gs,vrdoh,guo2025pgc,li2025wonderplay}, avatars~\cite{anigaussiananimatablegaussianavatar,gaussianavatar,gaussianavatars}, and embodied systems~\cite{lu2024manigaussian,shorinwa2024splatmovermultistageopenvocabularyrobotic,ji2024graspsplats,zheng2024gaussiangrasper,yu2025real2render2realscalingrobotdata}.

Existing deformable GS methods largely follow a reconstruct-then-drive \\pipeline~\cite{PhysGaussian,mani-gs,GaussianMesh,sugar,physics3d,dreamphysics}: a static reconstruction is first obtained from multi-view images, then Gaussians are driven by particle-based or mesh-based deformation. This design introduces two fundamental limitations. First, appearance is typically baked into per-Gaussian colors, yielding physically inconsistent responses under deformation and relighting, especially for glossy surfaces. Second, the static reconstruction stage restricts post-hoc editing of materials. While inverse rendering can disentangle geometry, materials, and illumination for photorealistic relighting, existing solutions remain largely confined to static scenes.

We present \textit{\textbf{\modelname{}}}, a unified deformable Gaussian framework that jointly supports inverse rendering, relighting, physics-based simulation, and 3D animation within a shared representation. \modelname{} models materials from diffuse to specular for reconstruction, and introduces a hybrid driving mechanism that supports both particle systems and mesh deformations. Following the separation principle in GSP~\cite{gsp}, we decouple simulation from rendering and update Gaussians through generalized moving least squares (GMLS) interpolation~\cite{GMLS}.

To render dynamic scenes under changing illumination, \modelname{} estimates the full rendering equation with low-sample Monte Carlo integration and multiple importance sampling (MIS), combining cosine-weighted, GGX~\cite{Heitz2018GGX}, and environment-light sampling~\cite{PBR}, followed by cross-bilateral denoising~\cite{SVGF}. For efficiency, we store decoupled material parameters on TSDF-extracted mesh vertices and perform mesh-based ray tracing, replacing 2D Gaussian ray tracing~\cite{IRGS} during continuous deformation. This design preserves high-quality shadows and inter-reflections while significantly improving runtime.

To summarize, our contributions include: \textbf{(i)}\textbf{Unified deformable GS:} \modelname{} integrates physically-based inverse rendering, relighting, and deformation-driven control in a single Gaussian framework;
\textbf{(ii)}\textbf{Efficient dynamic rendering:} We combine MIS-based Monte Carlo rendering with denoising and mesh-based ray tracing to accelerate relighting under deformation efficiently;
\textbf{(iii)}\textbf{Editable asset pipeline:} A decoupled geometry--illumination--material \\pipeline enables high-fidelity creation and post-editing of 3D assets for content production and simulation.

\begin{figure}[t]
    \centering
    \includegraphics[width=1\linewidth]{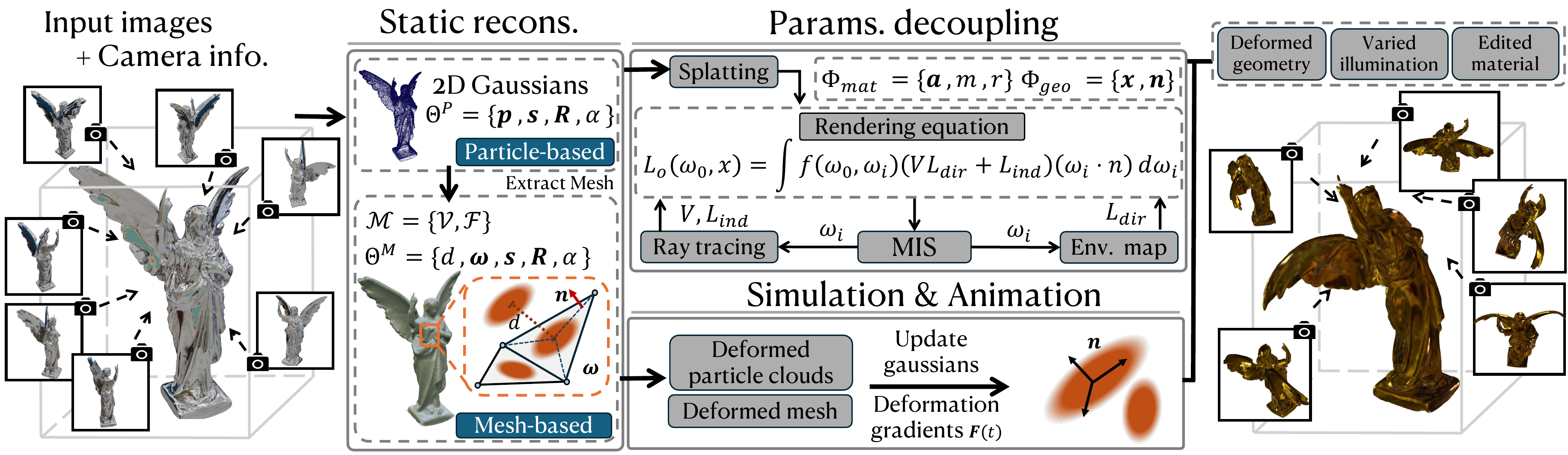} 
    \caption{
        \textbf{Overview of \textit{\modelname{}}.} Our framework consists of three stages:
        (i) Static reconstruction: building initial 2D Gaussians and mesh from calibrated images, with optional mesh-based reinitialization;
        (ii) Parameters decoupling: generating material (\textit{albedo, metallic, roughness}) and geometry (\textit{position, normal}) maps via splatting while solving the rendering equation through MIS and ray tracing;
        (iii) Dynamic driving: updating Gaussians via deformation gradients between canonical and deformed particle clouds/meshes from physical simulation/animation. Overall, our decoupled geometry-illumination-material framework supports physically plausible rendering under deformations, illumination changes, and material edits with interactive manipulation.
    }
    \label{fig:pipeline}
\end{figure}

\section{Related works}
\textbf{Dynamic Gaussian splatting.}
Prior work mainly falls into two paradigms. Learning-based methods take videos as input and encode temporal variation directly into Gaussians~\cite{deformable3dgs,pvg,4DGS,SpacetimeGS,gaussianavatars,gaussianavatar,yang2023gs4d,anigaussiananimatablegaussianavatar,BezierGS}. Deformation-driven methods follow a reconstruct-then-drive pipeline: they first reconstruct a static GS from multi-view images and then drive deformation via physics simulation or mesh manipulation. Our framework adopts the latter. Within this line, physics-inspired approaches couple GS with material-aware simulation~\cite{NeuMA,physdreamer,dreamphysics,physics3d,efficient,featuresplatting,vrdoh}, e.g., MPM in PhysGaussian~\cite{PhysGaussian} and PBD in GSP~\cite{gsp}, while geometry-driven methods rely on mesh editing for artist-controlled deformation~\cite{mani-gs,GaussianMesh,sugar,games,vr-gs}. Although effective for rendering static appearance, existing methods largely bake lighting into colors and thus cannot faithfully reproduce appearance changes under deformation or varying illumination, producing noticeable artifacts, especially on glossy surfaces. \modelname{} addresses this limitation with a decoupled representation that stores editable material parameters separately and renders them with physically-based rendering, improving realism in dynamic scenarios.

\noindent\textbf{Physically-based inverse rendering} aims to recover geometry, materials, and illumination from images, but is inherently ill-posed due to ambiguities between observations and physical parameters. Most methods rely on differentiable rendering with physics-based light transport. NeRF-based approaches~\cite{nerv,nerd,NeILF++,SAMURAI,nero,tensoflow,zhu2024multitimesmontecarlorendering} can model global illumination via ray marching over neural fields, but remain computationally expensive. In contrast, 3DGS~\cite{3DGS} provides an efficient explicit representation and has been extended with material modeling for inverse rendering~\cite{Gs-ir,R3DG,glossygs2024,PRTGS2024,GIR,SVG-IR,GI-GS,ref-gaussian}. Ref-Gaussian~\cite{ref-gaussian} uses deferred BRDF shading and mesh ray tracing for specular visibility, but the split-sum approximation can compromise material and lighting accuracy. IRGS~\cite{IRGS} evaluates the full rendering equation with 2D Gaussian ray tracing to capture inter-reflections, but its heavy stratified sampling is inefficient and often struggles on glossy surfaces. To support deformable objects under changing illumination, \modelname{} builds on these advances while explicitly maintaining the full physically-based rendering equation and improving efficiency in dynamic settings.

\section{Method}
This section introduces DR-GS, a decoupled deformable and relightable GS framework for physically-plausible rendering of deformable objects under varying illumination. Our three-stage pipeline first establishes the foundations of 2D Gaussian splatting and Gaussian ray tracing (Sec.~\ref{method:preliminary}). The static reconstruction stage builds initial 2D Gaussians and mesh representations from multi-view imagery (Sec.~\ref{method:static_recons}). We then solve the rendering equation to decouple material attributes from illumination while simultaneously  optimizing geometric parameters (Sec.~\ref{method:params_decoupling}). Finally, we develop a physically-based deformation-driven method for 2D Gaussians utilizing physical simulation and animation (Sec.~\ref{method:dynamic_driving}). The full pipeline is illustrated in Fig.~\ref{fig:pipeline}.

\subsection{Preliminary}
\label{method:preliminary}
\textbf{2D Gaussian splatting}~\cite{2DGS} addresses key limitations of 3DGS~\cite{3DGS}, including its lack of explicit surface normals and multi-view inconsistencies, through geometrically constrained explicit surface representation. It compresses 3D Gaussians into 2D surfels parameterized by a center $\bm{p}$, opacity $\alpha$, view-dependent color $\bm{c}$, axial scaling $\bm{s} = (s_u, s_v)$, and a rotation matrix $\bm{R}$ built from orthonormal tangents $\mathbf{t}_u$ and $\mathbf{t}_v$. The surface normal is explicitly derived as $\bm{n} = \bm{t}_u \times \bm{t}_v$, ensuring view-consistent geometry. A central innovation is ray-splat intersection, which maps screen-space pixels to UV coordinates via the Gaussian kernel $G(\bm{u}) = \exp\left(-(u^2 + v^2)/2\right)$, augmented with perspective-correct splatting to reduce multi-view artifacts. Rendering uses depth-ordered alpha blending:
\begin{equation}
\bm{c}(\bm{r}) = \sum_{i=1}^N \bm{c}_i \alpha_i \hat{G}_i(\bm{u}) \prod_{j=1}^{i-1} \left(1 - \alpha_j \hat{G}_j(\bm{u})\right)
\label{eq:alpha_blend},
\end{equation}
enabling end-to-end optimization of learnable parameters $\Theta_i = \{\bm{p}_i, \bm{s}_i, \bm{R}_i, \alpha_i, \bm{c}_i\}$.

\noindent\textbf{Gaussian ray tracing} integrates ray tracing techniques with Gaussian primitives to overcome limitations of rasterization-based Gaussian splatting in simulating effects like shadows and inter-reflections. The pioneering 3DGRT~\cite{3dgrt} introduces particle-based ray tracing for 3D Gaussians, using a $k$-buffer hit-based marching algorithm with OptiX~\cite{optix} hardware acceleration to improve both speed and accuracy. Following this advancement, 2DGRT~\cite{IRGS} resolves ray-splat inconsistencies through explicit surface representation and geometric constraints. This approach achieves physically accurate ray tracing, particularly for complex light paths involving multi-bounce indirect illumination.

\subsection{Static Reconstruction}
\label{method:static_recons}
\textbf{Initialization.} Our framework begins by pretraining with Ref-Gaussian~\cite{ref-gaussian} to acquire initial 2D Gaussians $\Theta^P=\{\bm{p}, \bm{s}, \bm{R}, \alpha\}$ and subsequently extracts triangular mesh $\mathcal{M}=\{\mathcal{V},\mathcal{F}\}$ via TSDF. DR-GS accommodates dual deformation approaches: particle-based deformation with optional mesh filling and mesh-based deformation necessitating reinitialization of $\Theta^P$ with $\mathcal{M}$.

\noindent\textbf{Mesh-based reinitialization.} We initialize a set of 2D Gaussians $\Theta^M=\{d, \bm{\omega}, \bm{s}, \bm{R}, \alpha\}$ on each triangular face $f\in \mathcal{F}$, where $\bm{R}$ aligns with the face normal $\bm{n}$, and $\bm{s}$ determined by local geometric properties. Same as GaussianMesh~\cite{GaussianMesh}, each Gaussian center $\bm{p}$ is parameterized by interpolation weights $\bm{\omega}=\{\omega_a, \omega_b, \omega_c\}$ and normal offset $d$, initialized as barycentric coordinates and zero, respectively. The relationship can be expressed as $\bm{p}=(\omega_a \bm{v}_a + \omega_b\bm{v}_b + \omega_c\bm{v}_c) + dr\bm{n}$, where $r$ denotes the circumradius of the associated triangle with vertices $\{\bm{v}_a,\bm{v}_b, \bm{v}_c\}$. Pretrained Gaussians $\Theta^P$ are projected onto mesh surfaces using spatial acceleration structure, retaining only interior projections. Finally, we compute averaged $\bm{\omega}$ and $\alpha$ per face and transfer them to corresponding mesh-based Gaussians $\Theta^M$, achieving geometry-aware reparameterization. See Alg.~\ref{alg:mesh_2dgs_init} for details.

\begin{algorithm}[H]
\caption{Mesh-based 2D Gaussian Initialization}
\label{alg:mesh_2dgs_init}
\textbf{Input:} Mesh $\mathcal{M}=(\mathcal{V},\mathcal{F})$, pretrained Gaussians $\Theta^P=\{(\boldsymbol{p}_i,\alpha_i)\}_{i=1}^{N_p}$. \\
\textbf{Output:} Face-attached Gaussians $\Theta^M=\{(d_j,\boldsymbol{\omega}_j,\boldsymbol{s}_j,\boldsymbol{R}_j,\alpha_j)\}_{j=1}^{N_f}$.
\begin{algorithmic}[1]
\State \textbf{Per-face init.} \For{$j=1,\dots,N_f$}
\State $\boldsymbol{n}_j \propto (\boldsymbol{v}_{j2}-\boldsymbol{v}_{j1})\times(\boldsymbol{v}_{j3}-\boldsymbol{v}_{j1})$, \ 
$\boldsymbol{R}_j \gets \mathrm{Align}(\boldsymbol{n}_j)$
\State $\boldsymbol{c}_j \gets (\boldsymbol{v}_{j1}+\boldsymbol{v}_{j2}+\boldsymbol{v}_{j3})/3$, \ 
$\tilde d_j \gets \min_{k\neq j}\|\boldsymbol{c}_j-\boldsymbol{c}_k\|$
\State $d_j\gets 0$, \ $\boldsymbol{\omega}_j\gets(1/3,1/3,1/3)$, \ $\boldsymbol{s}_j\gets \log(\tilde d_j+\epsilon)\boldsymbol{1}_2$
\EndFor

\State \textbf{Project to faces.} Build BVH on $\mathcal{M}$. \For{$i=1,\dots,N_p$}
\State $(d_i,f_i,\boldsymbol{\omega}_i)\gets \Pi_{\mathcal{M}}(\boldsymbol{p}_i)$ 
\If{$f_i\ge 0 \ \land\ \min(\boldsymbol{\omega}_i)>\tau$} 
    \State append $i$ to $\mathcal{I}_{f_i}$
\EndIf
\EndFor

\State \textbf{Aggregate \& transfer.} \For{$j=1,\dots,N_f$ with $\mathcal{I}_j\neq\emptyset$}
\State $\boldsymbol{\omega}_j \gets \frac{1}{|\mathcal{I}_j|}\sum_{i\in\mathcal{I}_j}\boldsymbol{\omega}_i$, \quad
$\alpha_j \gets \frac{1}{|\mathcal{I}_j|}\sum_{i\in\mathcal{I}_j}\alpha_i$
\State \textbf{(opt.)} $\forall i\in\mathcal{I}_j:\ (\boldsymbol{\omega}_i,\alpha_i)\gets(\boldsymbol{\omega}_j,\alpha_j)$
\EndFor
\end{algorithmic}
\end{algorithm}

\subsection{Parameters Decoupling}
\label{method:params_decoupling}

\textbf{Rasterization.} Following the same framework as IRGS~\cite{IRGS}, DR-GS employs a physically-based deferred rendering pipeline: Gaussians are first rasterized to generate per-pixel maps, after which the rendering equation is applied. Each Gaussian is augmented with a set of material parameters $\Phi_{mat}$, which includes albedo $\bm{a} \in [0,1]^3$, roughness $r \in [0,1]$, and metallic $m \in [0,1]$. Per-pixel attributes are aggregated via Gaussian rasterization:
\begin{equation}
\sum_{i=1}^{N} \gamma_i \{\bm{c}_i, d_i, \bm{n}_i, \bm{a}_i, r_i, m_i\}, \quad \text{where} \quad \gamma_i = \frac{T_i \alpha_i}{\sum_{k=1}^N T_k \alpha_k},\quad  T_{i} = \prod_{j=1}^{i-1} (1 - \alpha_{j}).
\label{eq:rasterization}
\end{equation}
Here, $\bm{c}_i$ denotes the outgoing radiance, $d_i$ represents depth, and $\bm{n}_i$ is the normal vector. Leveraging the resulting depth map, the 3D surface point $\bm{x}$ corresponding to each pixel can be computed.

\noindent\textbf{Physically-based rendering.} 
To achieve photorealistic rendering, we employ the complete rendering equation~\cite{rendering_equation}, which is the fundamental formulation simulating light transport in physically-based rendering (PBR):
\begin{equation}
\label{eq:rendering_equation}
L_o(\boldsymbol{\omega}_o, \boldsymbol{x}) = \int_{\Omega} f(\boldsymbol{\omega}_o, \boldsymbol{\omega}_i, \boldsymbol{x}) L_i(\boldsymbol{\omega}_i, \boldsymbol{x}) (\boldsymbol{\omega}_i \cdot \boldsymbol{n})  d\boldsymbol{\omega}_i,
\end{equation}
where $L_o$ and $L_i$ denote outgoing and incident radiance at point $\bm{x}$ in direction $\omega_o$ and $\omega_i$, respectively. The integral is computed over the hemisphere $\Omega$ around the surface normal $\bm{n}$. The bidirectional reflectance distribution function (BRDF) $f$ captures the material's light scattering behavior.

We further decompose incident light at the surface point $\boldsymbol{x}$ into direct and indirect components:
\begin{equation}
L_i(\boldsymbol{\omega}_i, \boldsymbol{x}) = V(\boldsymbol{\omega}_i, \boldsymbol{x}) L_{\text{dir}}(\boldsymbol{\omega}_i) + L_{\text{ind}}(\boldsymbol{\omega}_i, \boldsymbol{x}),
\end{equation}
where $L_{\text{dir}}$ represents distant illumination from an environment map, while visibility $V$ and indirect light $L_{\text{ind}}$ are calculated using 2DGRT~\cite{IRGS}. Note that $L_{\text{ind}}$ is handled differently during reconstruction training and dynamic driving stages: during training, it is obtained by alpha-blending outgoing radiance $\boldsymbol{c}_i$ from Gaussians; the relight stage methodology is detailed in Sec.~\ref{method:dynamic_driving}.

Given the incident radiance, we estimate the Eq.~\ref{eq:rendering_equation} via importance sampling~\cite{a_reflectance_model_for_computer_graphics}:
\begin{equation}
c_{\text{pbr}} = \frac{1}{N_r} \sum_{i=1}^{N_r} \frac{f(\boldsymbol{\omega}_o, \boldsymbol{\omega}_i, \boldsymbol{x}) L_i(\boldsymbol{\omega}_i, \boldsymbol{x}) (\boldsymbol{\omega}_i \cdot \boldsymbol{n})}{q(\boldsymbol{\omega}_i)},
\end{equation}
where $N_r$ directions $\boldsymbol{\omega}_i$ are drawn from proposal distribution $q$ with PDF $q(\boldsymbol{\omega}_i)$. 

\noindent\textbf{Training strategy.} Accurate modeling of light-surface interactions through geometry-aware ray tracing requires robust scene geometry. Precise disentanglement of geometry, illumination, and material is essential for flexible editing of virtual assets. We jointly optimize pretrained Gaussians $\Theta$ and material parameters $\Phi_{\text{mat}}$ while estimating illumination. To reduce computational cost, we selectively evaluate the Eq.~\ref{eq:rendering_equation} on a subset of pixels per view during training. The loss function is defined as:
\begin{equation}
\mathcal{L} = \mathcal{L}_c + \lambda_1^{\text{pbr}} \mathcal{L}_1^{\text{pbr}} + \lambda_{\text{light}} \mathcal{L}_{\text{light}},
\end{equation}
where $\mathcal{L}_c$ denotes the RGB reconstruction loss from 3DGS~\cite{3DGS} for rendered radiance $\mathcal{C}$, $\mathcal{L}_1^{\text{pbr}}$ represents the L$_1$ loss between physically-based rendered results and ground truth, and $\mathcal{L}_{\text{light}}$ regularizes incident illumination to natural white balance.

\subsection{Dynamic Driving}
\label{method:dynamic_driving}

\textbf{Particle-driven deformation}. Particle representations provide superior flexibility for modeling complex geometries. Conventional mesh extraction often yields non-manifold, self-intersecting, or non-watertight meshes~\cite{Flexicubes}, complicating physical simulations. High-resolution meshing incurs significant computational cost, while low-resolution alternatives lose geometric fidelity. We therefore employ particles for spatial discretization and leverage the MPM for simulation.

Specifically, we generate a particle cloud $\mathcal{P}$ by populating the interior of an extracted mesh $\mathcal{M}$ and incorporating its vertices. This hybrid representation supports subsequent mesh-based ray tracing and deformation-aware processing. Physical simulation is performed, yielding a deformed particle set. We then compute the deformation gradient $\bm{F}_i$ and updated center position $\bar{\bm{p}}_i$ for each 2D Gaussian via GMLS interpolation (detailed in Alg.~\ref{alg:gmls_interpolation}). The deformation gradient $\bm{F}_i$ is decomposed via polar decomposition into a rotation matrix $\bar{\bm{R}}_i$ and a scaling-shearing matrix $\bar{\bm{S}}_i$.
We efficiently apply $\bar{\bm{R}}_i$ and $\bar{\bm{S}}_i$ to each Gaussian as follows:
\begin{equation}
\label{eq:gaussian_update}
\bm{p}'_i = \bar{\bm{p}}_i, \quad 
\bm{R}'_i = \bar{\bm{R}}_i \bm{R}_i, \quad 
\bm{S}'_i = \text{diag}(\bar{\bm{\Lambda}_i}) \cdot \bm{S}_i,
\end{equation}
where
\begin{equation}
\label{eq:gaussian_update_}
\bar{\bm{\Lambda}_i} = 
\begin{bmatrix}
|\lambda_2(\bar{\bm{S}_i)}| & 0 \\
0 & |\lambda_1(\bar{\bm{S}_i})| \notag
\end{bmatrix}.
\end{equation}

\noindent\textbf{Mesh-driven deformation}. Mesh-based representations enable efficient editing, sculpting, animation, and relighting operations. For each deformed triangle $f' = (\bm{v_a}', \bm{v_b}', \bm{v_c}')$ in mesh $\mathcal{M}'$ and its corresponding face $f = (\bm{v_a}, \bm{v_b}, \bm{v_c})$ in the canonical mesh $\mathcal{M}$, we compute a rotation matrix $\bar{\mathbf{R}}_i$, a shearing matrix $\bar{\mathbf{S}}_i$, and face-based displacement, which are directly applied to the bound Gaussians.
\begin{equation}
\begin{aligned}
\Delta \mathbf{p} &= w_a (\bm{v}_a' - \bm{v}_a) + w_b (\bm{v}_b' - \bm{v}_b) + w_c (\bm{v}_c' - \bm{v}_c), \\
\bar{\mathbf{R}}_i &= w_a \bar{\mathbf{R}}_{v_a'} + w_b \bar{\mathbf{R}}_{v_b'} + w_c \bar{\mathbf{R}}_{v_c'}, \\
\bar{\mathbf{S}}_i &= w_a \bar{\mathbf{S}}_{v_a'} + w_b \bar{\mathbf{S}}_{v_b'} + w_c \bar{\mathbf{S}}_{v_c'}.
\end{aligned}
\end{equation}
The Gaussian center is updated as $\mathbf{p}'_i = \mathbf{p}_i + \Delta \mathbf{p}$, while the rotation and scale updates remain consistent with Eq. \ref{eq:gaussian_update} and Eq. \ref{eq:gaussian_update_}.

\noindent\textbf{Rendering acceleration and optimization}. During the dynamic driving phase, the original radiance values $\bm{c}_i$ are no longer valid due to changes in the illumination conditions from those in the static reconstruction stage. We need to aggregate material attributes via Gaussian ray tracing and estimate incident radiance. To address the computational bottleneck of ray tracing, we replace 2DGS-based tracing with mesh-based ray tracing in this stage. Material properties learned through physical-based parameter decoupling (detailed in Sec.~\ref{method:params_decoupling}) are stored on a triangular mesh extracted via TSDF, enabling accelerated attribute lookup at the first ray intersection.

To further improve ray sampling efficiency, we employ multiple importance sampling (MIS)~\cite{mis}, combining cosine-weighted, GGX, and environmental distributions to model diffuse, specular, and environmental lighting, respectively. The Monte Carlo estimator with the balance heuristic is defined as:
\begin{equation}
\sum_{i=1}^{n} \frac{1}{n_i} \sum_{j=1}^{n_i} w_i(X_{i,j}) \frac{g(X_{i,j})}{p_i(X_{i,j})}, \quad w_i(x) = \frac{n_i p_i(x)}{\sum_k n_k p_k(x)}.
\end{equation}
To mitigate inherent Monte Carlo noise, we incorporate a cross-bilateral filter based on spatiotemporal variance-guided filtering (SVGF)~\cite{SVGF}, which preserves geometric edges through depth- and normal-aware weighting.

\begin{algorithm}[t] 
\caption{GMLS Interpolation} \label{alg:gmls_interpolation} \footnotesize \textbf{Input:} Reference positions $\boldsymbol{X}_i \in \mathbb{R}^3$ , deformed positions $\boldsymbol{x}_i \in \mathbb{R}^3$, deformation gradients $\boldsymbol{F}_i \in \mathbb{R}^{3\times3}$, target points $\boldsymbol{y}_j \in \mathbb{R}^3$, binding matrix $\mathcal{B}_{jk}$ (maps target $j$ to source $k$), distance matrix $\mathcal{D}_{jk}$ \textbf{Output:} Interpolated positions $\boldsymbol{y}_j^{\text{interp}} \in \mathbb{R}^3$, interpolated deformation gradients $\boldsymbol{F}_j^{\text{interp}} \in \mathbb{R}^{3\times3}$ \begin{algorithmic}[1] \State \textbf{Initialize:} $\boldsymbol{y}_j^{\text{interp}} \gets \boldsymbol{0}$, $\boldsymbol{F}_j^{\text{interp}} \gets \boldsymbol{I}$ \For{each batch $\mathcal{J} \subseteq \{1,\ldots,N_g\}$} \State $\mathcal{N}_j \gets \{k | \mathcal{B}_{jk} = 1\}, \forall j \in \mathcal{J}$ \State $\boldsymbol{W}_j \gets \exp(-\mathcal{D}_{jk}^2/\sigma^2), k \in \mathcal{N}_j$ \State $\boldsymbol{\Phi}_k \gets [1, \boldsymbol{X}_k^T]^T$ \State $\boldsymbol{A}_j \gets \sum_{k\in\mathcal{N}_j} \boldsymbol{W}_j^k \boldsymbol{\Phi}_k \boldsymbol{\Phi}_k^T + \epsilon\boldsymbol{I}$ \State $\boldsymbol{b}_j \gets \sum_{k\in\mathcal{N}_j} \boldsymbol{W}_j^k \boldsymbol{\Phi}_k \boldsymbol{x}_k^T$ \State $\boldsymbol{\alpha}_j \gets \boldsymbol{A}_j^{-1}\boldsymbol{b}_j$ \State $\boldsymbol{y}_j^{\text{interp}} \gets \boldsymbol{\Phi}(\boldsymbol{0})^T \boldsymbol{\alpha}_j$ \State $\boldsymbol{F}_j^{\text{interp}} \gets \sum_{k\in\mathcal{N}_j} \boldsymbol{W}_j^k \boldsymbol{\Phi}_k \boldsymbol{F}_k \boldsymbol{A}_j^{-1} \boldsymbol{\Phi}(\boldsymbol{0})$ \EndFor \end{algorithmic} \end{algorithm}

\section{Experiment}
\label{sec:experiment}

\subsection{Experiment Setup}
\textbf{Datasets and metrics.}
To evaluate the proposed method, we conducted experiments on the widely-used GlossySynthetic dataset~\cite{nero}, TensoIR dataset~\cite{Jin2023TensoIR} and character models from Sketchfab and Mixamo (including Vegeta, Mutant, NotEnrique). For quantitative assessment of static reconstruction and relighting quality, we employed three metrics: PSNR, SSIM~\cite{ssim}, and LPIPS~\cite{LPIPS}. Since ground truth images are unavailable after deformation, quantitative evaluation is infeasible. We therefore conduct a user study with three perceptual criteria: (i) Physical plausibility (PP), assessing whether light transport appears physically realistic; (ii) Temporal coherence (TC), assessing flickering and temporal instability in dynamic sequences; and (iii) Deformation consistency (DC), assessing whether local appearance details remain consistent before and after deformation. We invite 30 participants to rate each criterion on a 0-5 scale and report the mean scores. The corresponding videos are provided in the project page.

\noindent\textbf{Baselines.}
We extensively compare with dynamic Gaussian splatting baselines (\textit{PhysGaussian}~\cite{PhysGaussian}, \textit{GSP}~\cite{gsp}, \textit{SuGaR}~\cite{sugar}, \textit{Mani-GS}~\cite{mani-gs}, \textit{GaussianMesh}~\cite{GaussianMesh}) and physics-based relightable reconstruction methods (\textit{GSP}~\cite{gsp}, \textit{IRGS}~\cite{IRGS}, \textit{Ref-Gaussian}~\cite{ref-gaussian}, GS-ROR$^2$~\cite{gs-ror2}). Notably, GSP serves as our primary baseline for simultaneously supporting deformation and relighting. For fair comparison, all methods are implemented using identical input data.

\noindent\textbf{Implementation details.}
We first strictly adhere to the original configuration of Ref-Gaussian~\cite{ref-gaussian} for pre-training, followed by an extended fine-tuning phase of 20,000 iterations. For MIS, 512 rays are sampled during reconstruction (256 cosine-weighted, 128 GGX, and 128 light samples), while the number is reduced to 32 rays (16 cosine-weighted, 8 GGX, and 8 light samples) for dynamic driving and relighting stage. Environment maps are set to a resolution of 128$\times$256.

\subsection{Evaluations and Comparisons}
\subsubsection{Qualitative comparison.}
\begin{figure}
    \centering
    \includegraphics[width=1\linewidth]{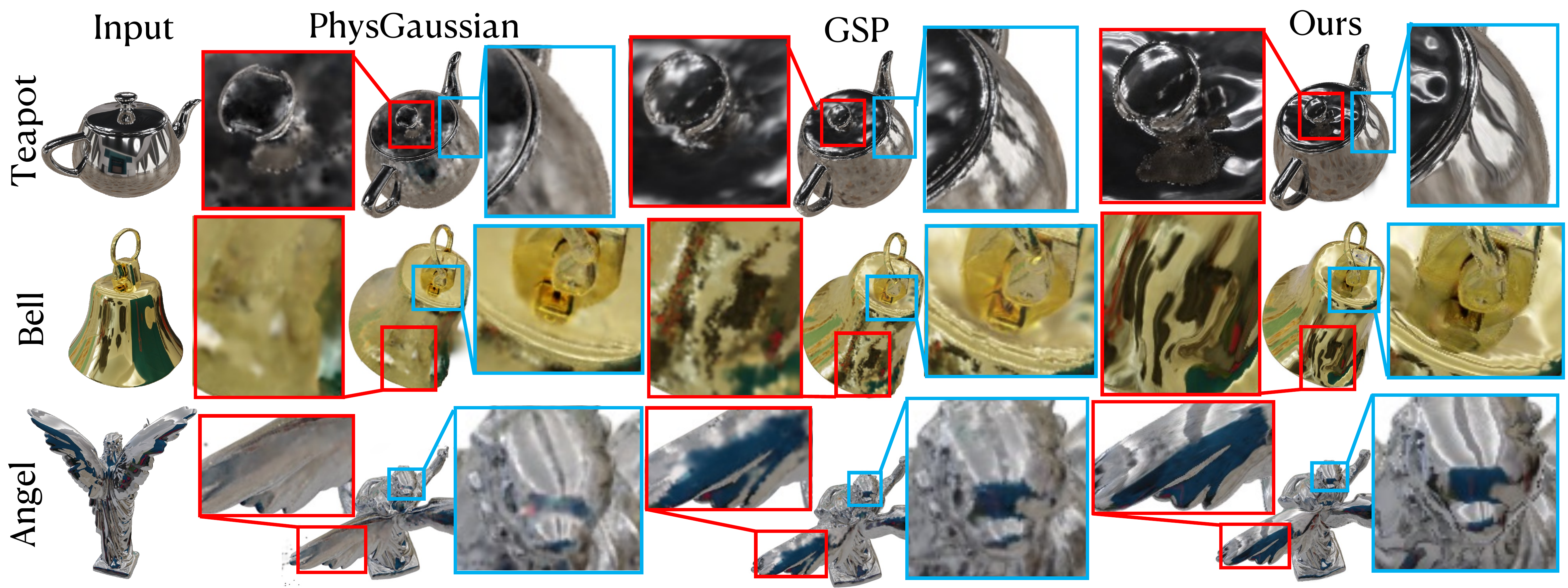} 
    \caption{
    \textbf{Qualitative comparisons on deformed reflective scenes}. Compared with PhysGaussian~\cite{PhysGaussian} and GSP~\cite{gsp}, DR-GS yields the clearest, most physically plausible glossy reflections under large deformations.
    }
    \label{fig:compare_deformed_reflective_scenes}
\end{figure}

Rendering results under geometric deformation are shown in Fig.~\ref{fig:compare_deformed_reflective_scenes}. PhysGaussian bakes environment reflections into static textures, causing severe artifacts on \textit{Angel} under motion, while GSP’s inaccurate surface reconstruction fails on glossy objects. In contrast, DR-GS preserves crisp, consistent metallic appearance even under large deformations.

\begin{figure}[htb]
    \centering
    \includegraphics[width=1\linewidth]{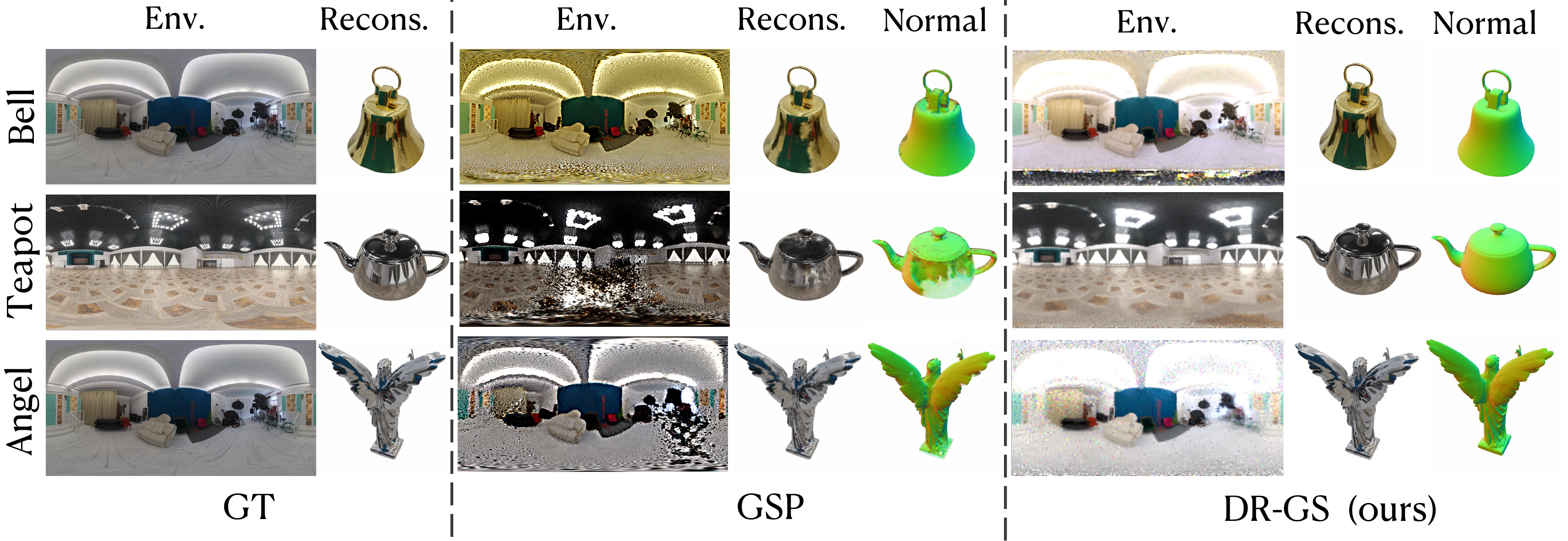} 
    \caption{
    \textbf{Qualitative comparison of illumination decomposition and reconstruction}. Compared with GSP, DR-GS better decouples environment lighting and yields clearer, more photorealistic novel-view renderings.
    }
    \label{fig:appendix_env_recons}
\end{figure}

\begin{figure}[tbp]
    \centering
    \begin{minipage}{0.66\linewidth}
        \centering
        \includegraphics[width=\linewidth]{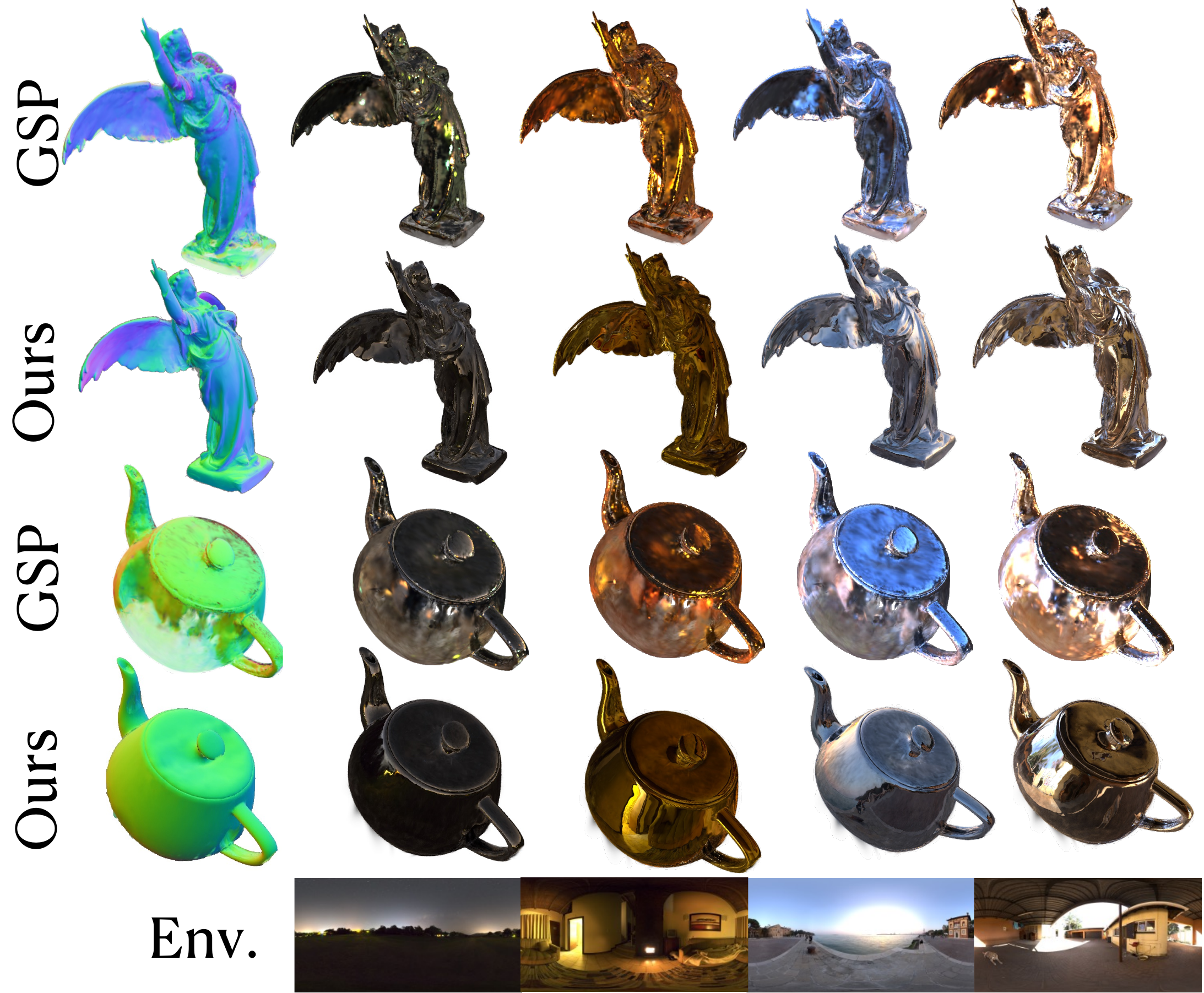}
        \caption{
        \textbf{Deformation results under relighting}. Under combined deformation and lighting changes, DR-GS yields smoother normals and physically plausible reflections.
        }
        \label{fig:compare_relight_deform}
    \end{minipage}
    \hspace{0.01\linewidth}
    \begin{minipage}{0.30\linewidth}
        \centering
        \includegraphics[width=0.9\linewidth]{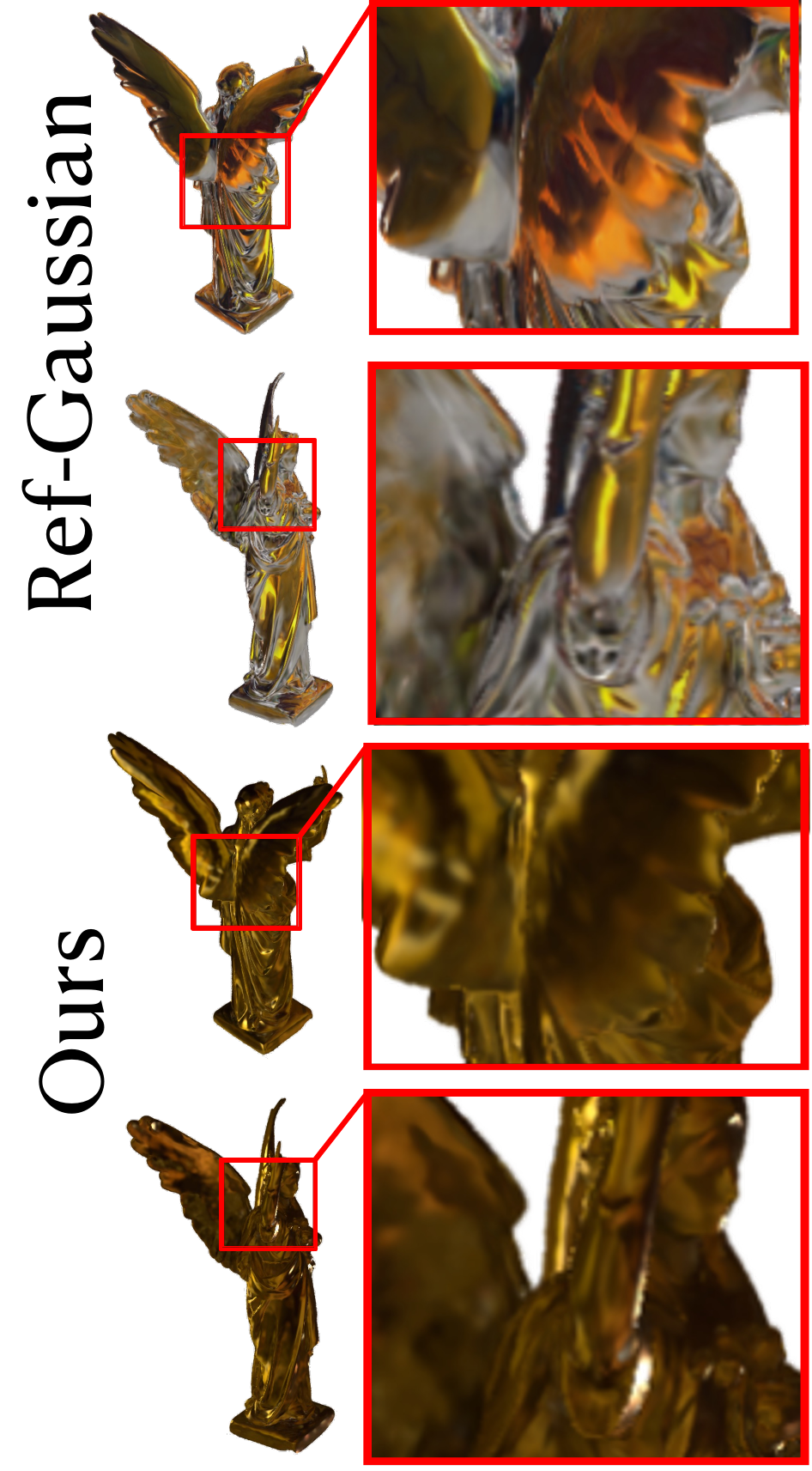}
        \caption{
        \textbf{Relighting under dim illumination}. Consistent metallic sheen suggests DR-GS learns a more accurate base color.}
        \label{fig:compare_refgs}
    \end{minipage}
\end{figure}

As further shown in Fig.~\ref{fig:appendix_env_recons} and Fig.~\ref{fig:compare_relight_deform}, GSP suffers from surface irregularities and holes that are amplified by deformation, leading to catastrophic relighting. DR-GS instead reconstructs smooth surfaces and remains physically plausible under joint deformation and illumination changes.

\begin{figure}
    \centering
    \includegraphics[width=1\linewidth]{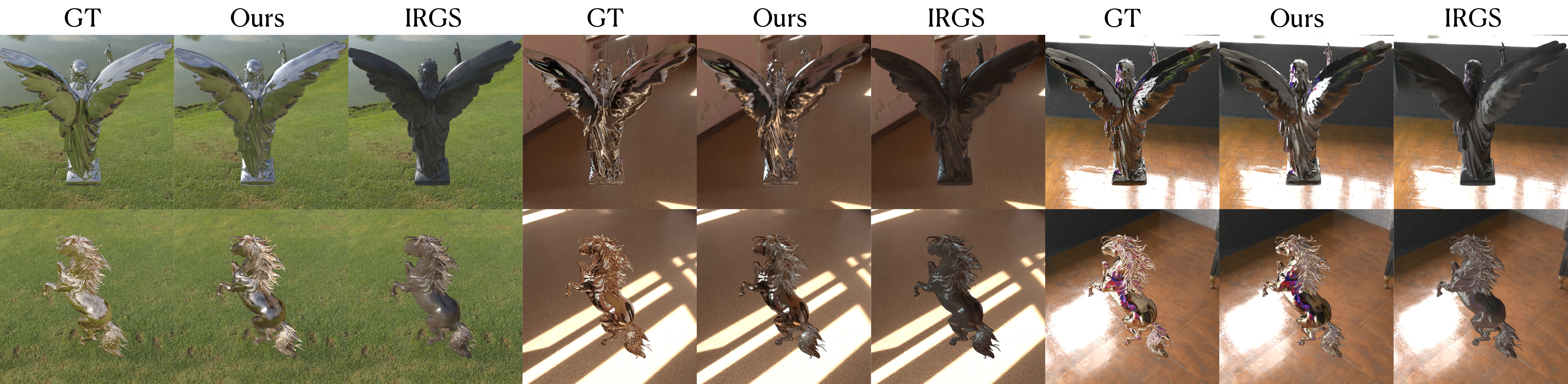} 

    \caption{
    \textbf{Relighting results on glossy objects}. DR-GS more faithfully matches the GT specular highlights and reflection details while preserving material consistency, whereas IRGS often appears darker with weakened reflections and lost fine details.
    }
    \label{fig:compare_irgs}
\end{figure}

Under novel illumination (Fig.~\ref{fig:compare_refgs}), Ref-Gaussian produces mottled relighting due to insufficient material–lighting disentanglement, whereas DR-GS stays spatially coherent. IRGS yields implausibly dim glossy relighting (Fig.~\ref{fig:compare_irgs}) due to missing importance sampling and metallic modeling.

\subsubsection{Quantitative comparison.}
Table~\ref{tb:quant_main_compare_detail} and Table~\ref{tb:glossysyn_relit} summarize quantitative results on the GlossySynthetic dataset, where darker red indicates better performance. Table~\ref{tb:quant_main_compare_detail} compares per-scene static reconstruction quality across dynamic methods, while Table~\ref{tb:glossysyn_relit} evaluates relighting quality and additionally reports efficiency in terms of training time and rendering speed. Overall, our method achieves the best or competitive reconstruction and relighting performance across most scenes, while maintaining favorable computational efficiency.

Due to the lack of ground-truth after dynamic deformation, we evaluate visual quality via a user study (Table~\ref{tb:user_study}). Our method achieves the highest ratings across all metrics, and remains superior under relighting. GSP degrades noticeably after relighting, while PhysGaussian does not support relighting and is therefore marked as N/A.

\subsection{Ablations}
\begin{figure}[tbp]
    \centering
    \begin{minipage}{0.36\linewidth}
        \centering
        \includegraphics[width=\linewidth]{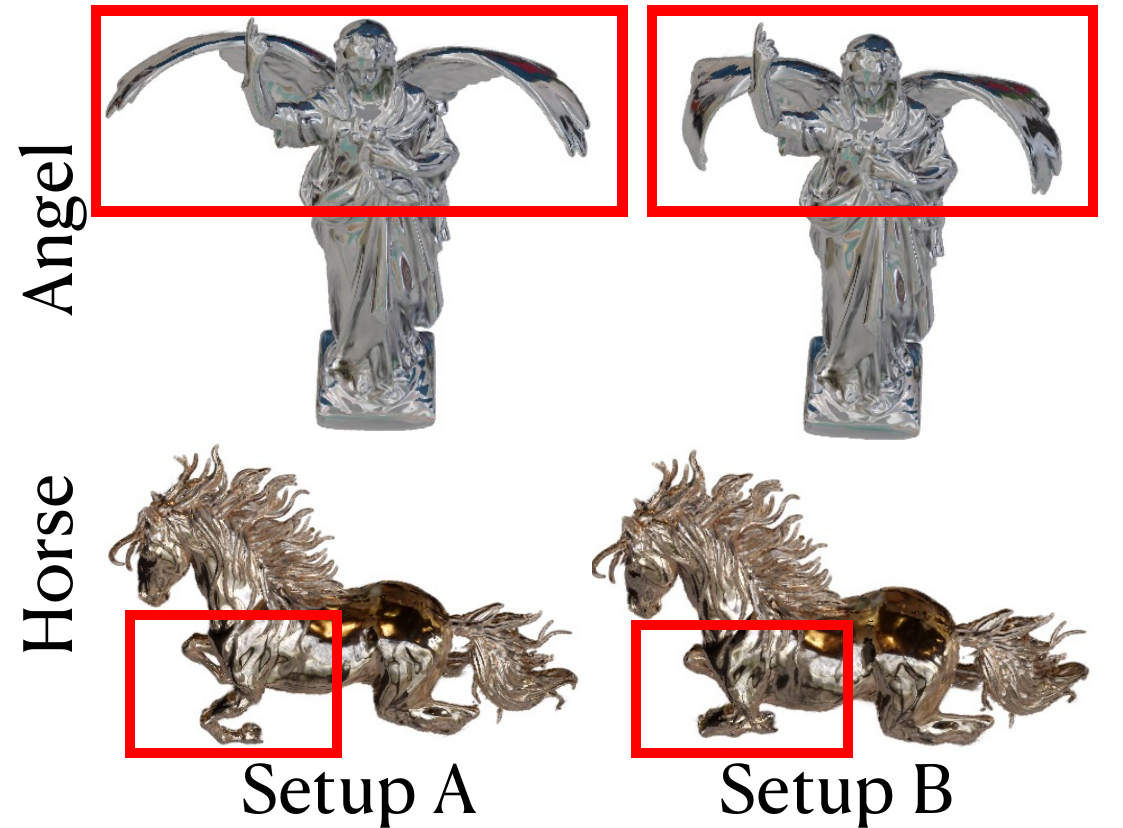}
        \caption{
        \textbf{Simulated deformation comparison}. The higher Young’s modulus in Setup A diminished its elastic deformability compared to Setup B.
        }
        \label{fig:more_exp_edited_sim}
    \end{minipage}
    \hspace{0.01\linewidth}
    \begin{minipage}{0.60\linewidth}
        \centering
        \includegraphics[width=\linewidth]{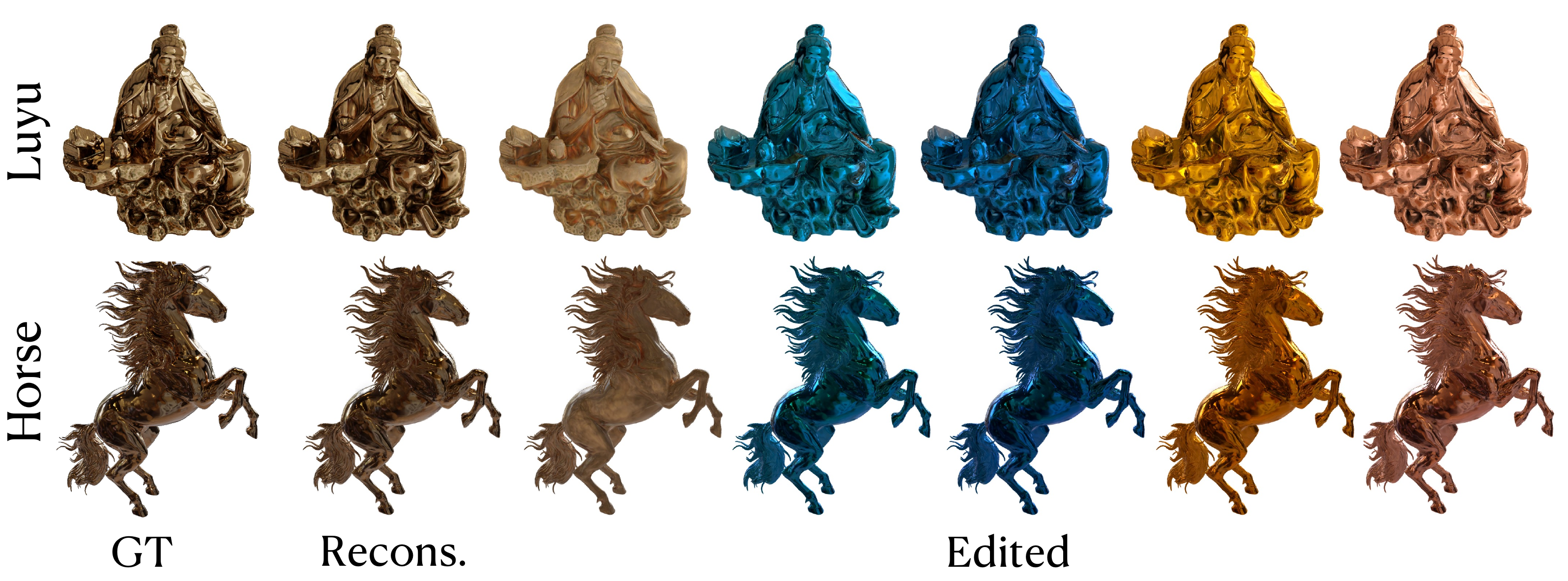}
        \caption{
        \textbf{PBR results with material parameter editing}. We present a comparison between DR-GS static reconstruction results and ground truth, along with rendered results after editing the decoupled material parameters.
        }
        \label{fig:more_exp_edited_render}
    \end{minipage}
\end{figure}

\noindent\textbf{Parameter editing.}
DR-GS’s fully decoupled parameters enable flexible post-reconstruction editing of digital assets. Fig.~\ref{fig:more_exp_edited_sim} and Fig.~\ref{fig:more_exp_edited_render} showcase edits for simulation and rendering: the former demonstrates gravity-driven soft-body deformation, where parameter changes markedly affect the Angel’s wings and the Horse’s legs (red boxes), while the latter shows that joint PBR edits produce diverse and photorealistic appearances.

\begin{figure}
    \centering
    \begin{minipage}{0.35\linewidth}
        \centering
        \includegraphics[width=\linewidth]{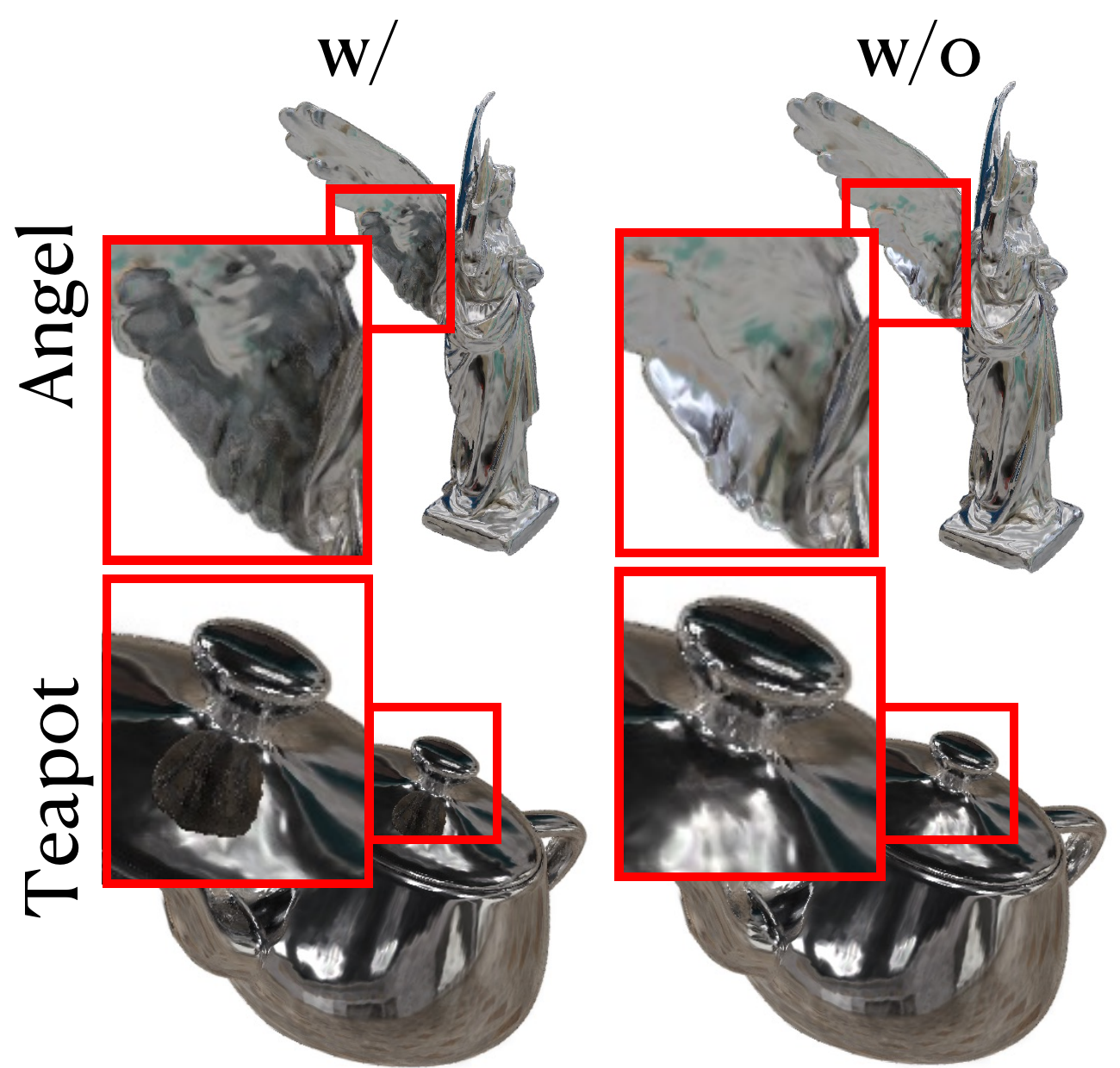}

        \caption{ 
        \textbf{Ablation on modeling of inter-reflection}.
        }
        \label{fig:ab_inter_reflection}
    \end{minipage}
    \hspace{0.01\linewidth}
    \begin{minipage}{0.62\linewidth}
        \centering
        \includegraphics[width=\linewidth]{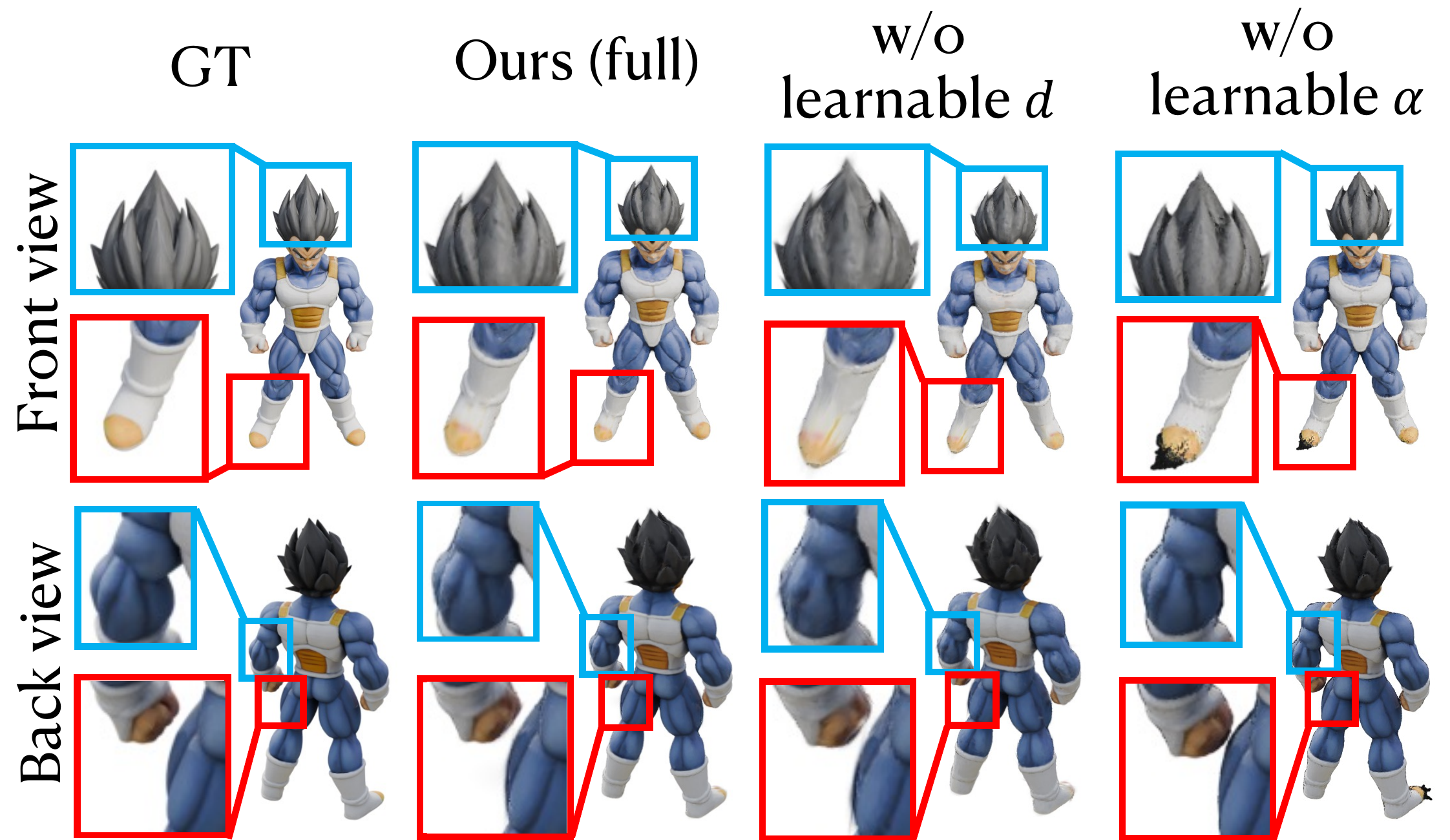}
        \caption{
        \textbf{Ablation on learnable parameters}. 
        }
        \label{fig:ab_learnable_params}
    \end{minipage}
\end{figure}

\noindent\textbf{Gaussian-based inter-reflection.}
Reflective surfaces can generate indirect illumination through their own geometric structures, leading to inter-reflection effects. As shown in Fig.~\ref{fig:ab_inter_reflection}, DR-GS incorporates ray-traced visibility to synthesize plausible indirect illumination under novel views, achieving physically faithful modeling of inter-reflection. 

\begin{table}
\caption{Per-scene static reconstruction quality comparison on GlossySynthetic~\cite{nero} dataset across dynamic methods. 
The intensity of the red color signifies a better result. \textbf{(P)} and \textbf{(M)} denote particle-driven and mesh-driven variants, respectively.}
\label{tb:quant_main_compare_detail}
\centering
\scriptsize
\renewcommand{\arraystretch}{1.1}
\setlength{\tabcolsep}{1.2mm}
\begin{adjustbox}{max width=\linewidth}
\begin{tabular}{@{}c|ccc|ccc|ccc|ccc}
\hline\hline

\textbf{Scenes} & \multicolumn{3}{c|}{\textbf{angel}} & \multicolumn{3}{c|}{\textbf{bell}} & \multicolumn{3}{c|}{\textbf{cat}} & \multicolumn{3}{c}{\textbf{horse}} \\
\hline
\textbf{Metrics} & \textbf{PSNR\textuparrow} & \textbf{SSIM\textuparrow} & \textbf{LPIPS\textdownarrow} & \textbf{PSNR\textuparrow} & \textbf{SSIM\textuparrow} & \textbf{LPIPS\textdownarrow} & \textbf{PSNR\textuparrow} & \textbf{SSIM\textuparrow} & \textbf{LPIPS\textdownarrow} & \textbf{PSNR\textuparrow} & \textbf{SSIM\textuparrow} & \textbf{LPIPS\textdownarrow} \\ 
\hline
PhysGaussian(P) & 25.38 & 0.801 & 0.092 & 24.31 & 0.904 & 0.122 & 29.48 & 0.947 & 0.085 & 24.49 & 0.790 & 0.087 \\ 
\hline
GSP(P) & 25.70 & \cellcolor{red!30}0.916 & 0.083 & \cellcolor{red!30}29.26 & \cellcolor{red!30}0.940 & \cellcolor{red!30}0.090 & \cellcolor{red!30}30.82 & \cellcolor{red!30}0.955 & \cellcolor{red!30}0.068 & \cellcolor{red!50}25.90 & \cellcolor{red!30}0.926 & \cellcolor{red!30}0.064 \\ 
\hline
SuGaR(M) & 12.81 & 0.775 & 0.202 & 12.40 & 0.740 & 0.240 & 12.51 & 0.730 & 0.203 & 14.31 & 0.748 & 0.228 \\ 
\hline
Mani-GS(M) & \cellcolor{red!30}26.39 & \cellcolor{red!10}0.911 & \cellcolor{red!30}0.075 & 24.58 & 0.901 & 0.114 & \cellcolor{red!10}29.90 & \cellcolor{red!10}0.950 & \cellcolor{red!10}0.072 & \cellcolor{red!10}24.76 & \cellcolor{red!10}0.897 & \cellcolor{red!10}0.075 \\ 
\hline
GaussianMesh(M) &  \cellcolor{red!10}25.84 &  0.901 & \cellcolor{red!10}0.079 & 24.02 & 0.890 & 0.118 & 28.39 & 0.936 & 0.083 & 23.70 & 0.886 & 0.081 \\ 
\hline
\textbf{ours(P)} & \cellcolor{red!50}29.64 & \cellcolor{red!50}0.943 & \cellcolor{red!50}0.052 & \cellcolor{red!50}31.94 & \cellcolor{red!50}0.962 & \cellcolor{red!50}0.056 & \cellcolor{red!50}31.25 & \cellcolor{red!50}0.965 & \cellcolor{red!50}0.052 & \cellcolor{red!30}25.54 & \cellcolor{red!50}0.929 & \cellcolor{red!50}0.055 \\ 
\hline
\textbf{ours(M)} & 25.04 & 0.887 & 0.094 & \cellcolor{red!10}28.12 & \cellcolor{red!10}0.917 & \cellcolor{red!10}0.103 & 27.07 & 0.931 & 0.091 & 21.25 & 0.851 & 0.111 \\ 
\hline\hline
\textbf{Scenes} & \multicolumn{3}{c|}{\textbf{luyu}} & \multicolumn{3}{c|}{\textbf{potion}} & \multicolumn{3}{c|}{\textbf{tbell}} & \multicolumn{3}{c}{\textbf{teapot}} \\ 
\hline
\textbf{Metrics} & \textbf{PSNR\textuparrow} & \textbf{SSIM\textuparrow} & \textbf{LPIPS\textdownarrow} & \textbf{PSNR\textuparrow} & \textbf{SSIM\textuparrow} & \textbf{LPIPS\textdownarrow} & \textbf{PSNR\textuparrow} & \textbf{SSIM\textuparrow} & \textbf{LPIPS\textdownarrow} & \textbf{PSNR\textuparrow} & \textbf{SSIM\textuparrow} & \textbf{LPIPS\textdownarrow} \\ 
\hline
PhysGaussian(P) & 26.22 & 0.904 & 0.086 & \cellcolor{red!30}28.71 & \cellcolor{red!10}0.923 & 0.132 & 22.76 & \cellcolor{red!10}0.893 & 0.153 & 21.14 & 0.877 & 0.122 \\ 
\hline
GSP(P) & \cellcolor{red!30}27.19 & \cellcolor{red!30}0.916 & 0.073 & \cellcolor{red!10}28.41 & \cellcolor{red!30}0.932 & 0.113 & \cellcolor{red!30}24.52 & \cellcolor{red!30}0.910 & \cellcolor{red!30}0.131 & \cellcolor{red!10}22.67 & \cellcolor{red!30}0.898 & 0.108 \\ 
\hline
SuGaR(M) & 15.15 & 0.754 & 0.216 & 14.68 & 0.613 & 0.254 & 12.77 & 0.449 & 0.434 & 13.62 & 0.744 & 0.194 \\ 
\hline
Mani-GS(M) & \cellcolor{red!10}26.42 & \cellcolor{red!10}0.909 & \cellcolor{red!30}0.069 & 27.83 & 0.922 & \cellcolor{red!30}0.110 & 21.38 & 0.885 & \cellcolor{red!10}0.145 & 21.18 & 0.878 &  \cellcolor{red!30}0.106 \\ 
\hline
GaussianMesh(M) & 26.02 & 0.905 & \cellcolor{red!10}0.071 & 27.53 & 0.918 & \cellcolor{red!10}0.112 & 20.55 & 0.871 & 0.156 & 20.64 & 0.871 & 0.109 \\ 
\hline
\textbf{ours(P)} & \cellcolor{red!50}28.57 & \cellcolor{red!50}0.940 & \cellcolor{red!50}0.053 & \cellcolor{red!50}30.55 & \cellcolor{red!50}0.941 & \cellcolor{red!50}0.095 & \cellcolor{red!50}27.54 & \cellcolor{red!50}0.942 & \cellcolor{red!50}0.094 & \cellcolor{red!50}26.72 & \cellcolor{red!50}0.945 & \cellcolor{red!50}0.062 \\ 
\hline
\textbf{ours(M)} & 24.10 & 0.873 & 0.098 & 26.75 & 0.887 & 0.145 & \cellcolor{red!10}23.67 & 0.879 & 0.165 &  \cellcolor{red!30}22.93 &  \cellcolor{red!10}0.882 &  \cellcolor{red!10}0.107 \\ 
\hline\hline

\end{tabular}
\end{adjustbox}
\end{table}

\begin{table}
\caption{Quantitative relighting comparison on the GlossySynthetic~\cite{nero} dataset. The intensity of the red color signifies a better result.}
\label{tb:glossysyn_relit}
\centering
\scriptsize
\renewcommand{\arraystretch}{1.1}
\setlength{\tabcolsep}{1.2mm}
\begin{adjustbox}{max width=\linewidth}
\begin{tabular}{@{}c|cc|cc|cc|cc|c}
\hline\hline
\textbf{Scenes}
& \multicolumn{2}{c|}{\textbf{angel}}
& \multicolumn{2}{c|}{\textbf{bell}}
& \multicolumn{2}{c|}{\textbf{cat}}
& \multicolumn{2}{c|}{\textbf{horse}}
& \textbf{Training time}

\\
\hline
\textbf{Metrics}
& \textbf{PSNR$\uparrow$} & \textbf{SSIM$\uparrow$}
& \textbf{PSNR$\uparrow$} & \textbf{SSIM$\uparrow$}
& \textbf{PSNR$\uparrow$} & \textbf{SSIM$\uparrow$}
& \textbf{PSNR$\uparrow$} & \textbf{SSIM$\uparrow$}
& \textbf{Hour}
\\

\hline 
GS-ROR$^2$ & 20.81 & 0.878 & \cellcolor{red!30}24.49 & \cellcolor{red!30}0.927 & \cellcolor{red!30}26.28 & \cellcolor{red!50}0.942 & 23.31 & 0.938 & 1.5 \\

\hline
Ref-Gaussian & \cellcolor{red!30}21.39 & \cellcolor{red!30}0.900 & 22.90 & 0.919 & 20.54 & 0.912 & \cellcolor{red!50}24.97 & \cellcolor{red!50}0.944 & 0.58 \\
\hline
IRGS & 20.58 & 0.860 & 20.98 & 0.878 & 22.43 & 0.888 & 22.10 & 0.921 & 1.0 \\
\hline 
\textbf{ours} & \cellcolor{red!50}24.15 & \cellcolor{red!50}0.905 & \cellcolor{red!50}25.81 & \cellcolor{red!50}0.931 & \cellcolor{red!50}27.34 & \cellcolor{red!30}0.940 & \cellcolor{red!30}24.19 & \cellcolor{red!30}0.939 & 0.68
 \\

\hline\hline
\textbf{Scenes} & \multicolumn{2}{c|}{\textbf{luyu}} & \multicolumn{2}{c|}{\textbf{potion}} & \multicolumn{2}{c|}{\textbf{tbell}} & \multicolumn{2}{c|}{\textbf{teapot}} & \textbf{Render time} \\ 
\hline
\textbf{Metrics}
& \textbf{PSNR$\uparrow$} & \textbf{SSIM$\uparrow$}
& \textbf{PSNR$\uparrow$} & \textbf{SSIM$\uparrow$}
& \textbf{PSNR$\uparrow$} & \textbf{SSIM$\uparrow$}
& \textbf{PSNR$\uparrow$} & \textbf{SSIM$\uparrow$}
& \textbf{FPS}
\\
\hline
GS-ROR$^2$ & 22.61 & \cellcolor{red!50}0.899 & \cellcolor{red!30}25.67 & \cellcolor{red!30}0.918 & \cellcolor{red!50}22.80 & \cellcolor{red!50}0.918 & 21.17 & 0.893 & 122 \\

\hline
Ref-Gaussian & 19.74 & 0.875 & 20.06 & 0.868 & 20.74 & \cellcolor{red!30}0.904 & \cellcolor{red!30}21.78 & \cellcolor{red!30}0.924 & 208 \\
\hline
IRGS & \cellcolor{red!30}22.73 & 0.852 & 22.92 & 0.866 & 19.97 & 0.853 & 19.27 & 0.870 & 0.5 \\
\hline
\textbf{ours} & \cellcolor{red!50}25.16 & \cellcolor{red!30}0.889 & \cellcolor{red!50}27.35 & \cellcolor{red!50}0.923 & \cellcolor{red!30}22.12 & 0.884 & \cellcolor{red!50}23.08 & \cellcolor{red!50}0.925 & 24 \\
\hline\hline

\end{tabular}
\end{adjustbox}
\end{table}

\begin{table}
\caption{\textbf{User study}.}
\label{tb:user_study}

\newcolumntype{C}[1]{>{\centering\let\newline\\\arraybackslash\hspace{0pt}}m{#1}}
\begin{adjustbox}{max width=\linewidth}
\setlength{\tabcolsep}{1.2mm}
\begin{tabular}{@{} *{7}{C{2cm}} @{}} 
\hline\hline
\textbf{Metrics} & PhysGaussian & PhysGaussian(relight) &  GSP & GSP (relight) & \textbf{ours} & \textbf{ours (relight)} \\
\hline
\textbf{PP$\uparrow$} & 3.26 & N/A & 4.38 & 3.82 & \textbf{4.56} & \textbf{4.62} \\
\hline
\textbf{TC$\uparrow$} & 3.87 & N/A& 4.12 & 3.56 & \textbf{4.42} & \textbf{4.27} \\
\hline
\textbf{DC$\uparrow$} & 2.89 & N/A& 3.76 & 3.65 & \textbf{4.19} & \textbf{4.21} \\
\hline\hline
\end{tabular}
\end{adjustbox}
\end{table}

\begin{figure}[tbp]
    \centering
    \includegraphics[width=1\linewidth]{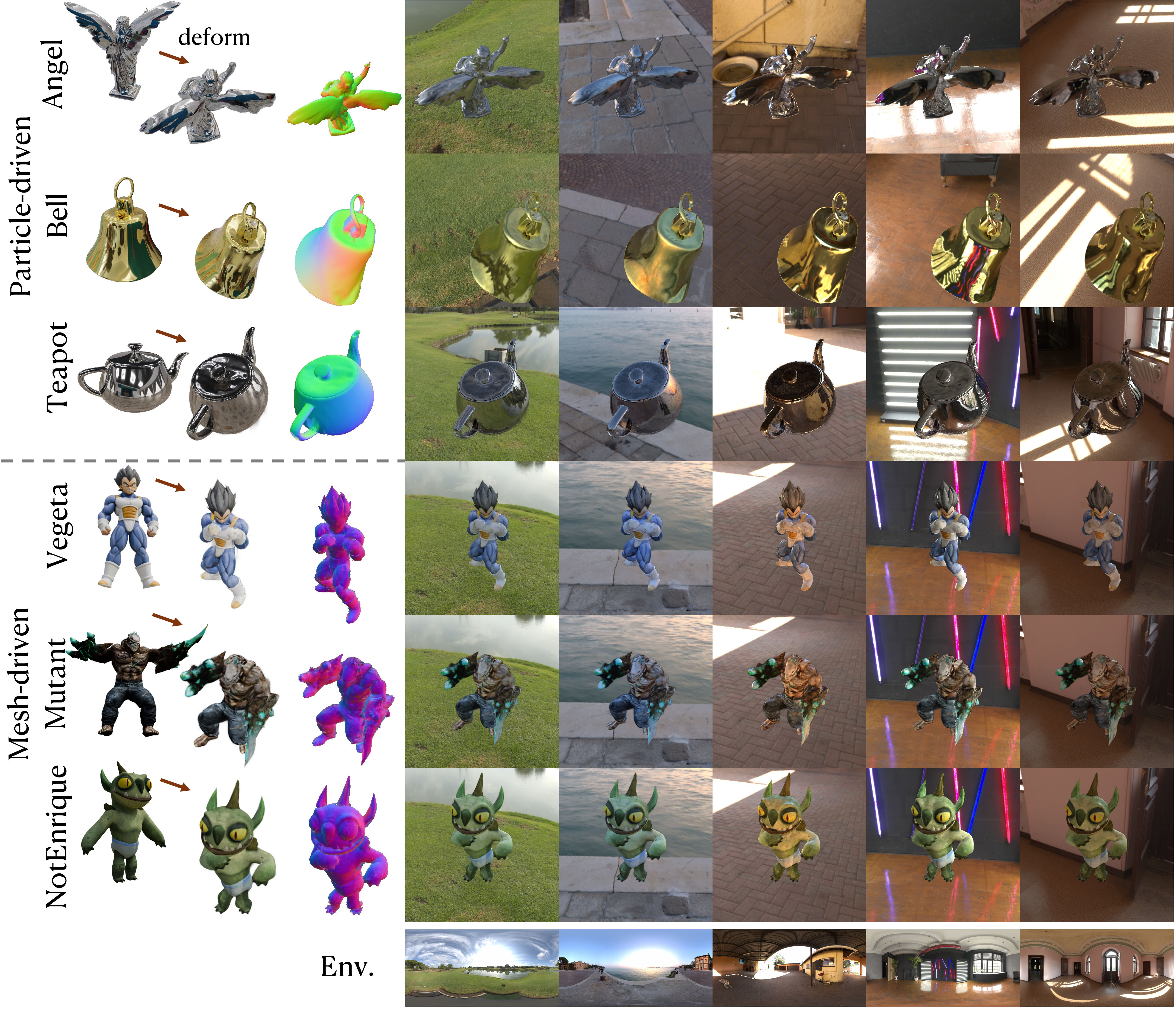} 

    \caption{
    \textbf{More results}. It demonstrates reconstruction, geometric deformation, and PBR under various illumination conditions using estimated material parameters. 
    }
    \label{fig:main_relight_deform}
\end{figure}

\noindent\textbf{Robustness to extracted mesh.}
Fig.~\ref{fig:ab_learnable_params} demonstrates that introducing learnable Gaussian attributes, i.e. normal offset $d$ and opacity $\alpha$, effectively enhances adaptability and robustness to inaccuracies in the extracted mesh. Without learnable $d$, the result exhibits noticeable spiky artifacts and structural blur; without learnable $\alpha$, scattered stain-like noise and localized crack-like distortions appear. In contrast, our full model achieves high-fidelity, clean, and detail-preserving rendering.

\begin{figure}[tbp]
    \centering
    \includegraphics[width=1\linewidth]{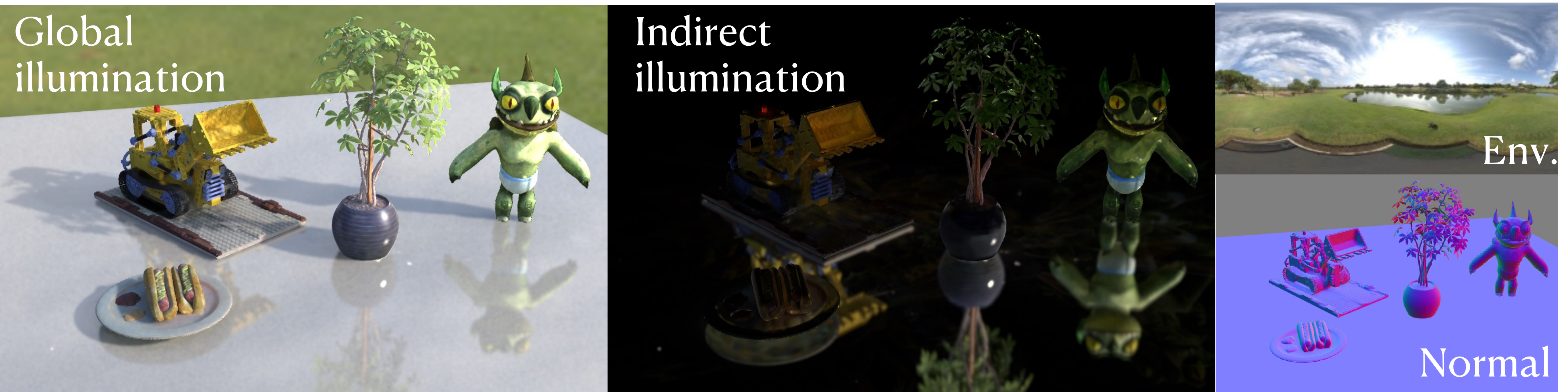} 

    \caption{
    \textbf{Composed scenes}. It demonstrates DR-GS’s ability to model occlusion-induced shadows and reflections arising from geometric visibility effects.
    }
    \label{fig:compose_shadow}
\end{figure}

\noindent\textbf{Shadow and reflection.}
Fig.~\ref{fig:compose_shadow} shows that DR-GS correctly models occlusion relationships induced by geometric visibility. The resulting shadows and reflections vary coherently with occlusions, yielding physically plausible occlusion-aware shading and reflections.

\begin{figure}
    \centering
    \includegraphics[width=0.75\linewidth]{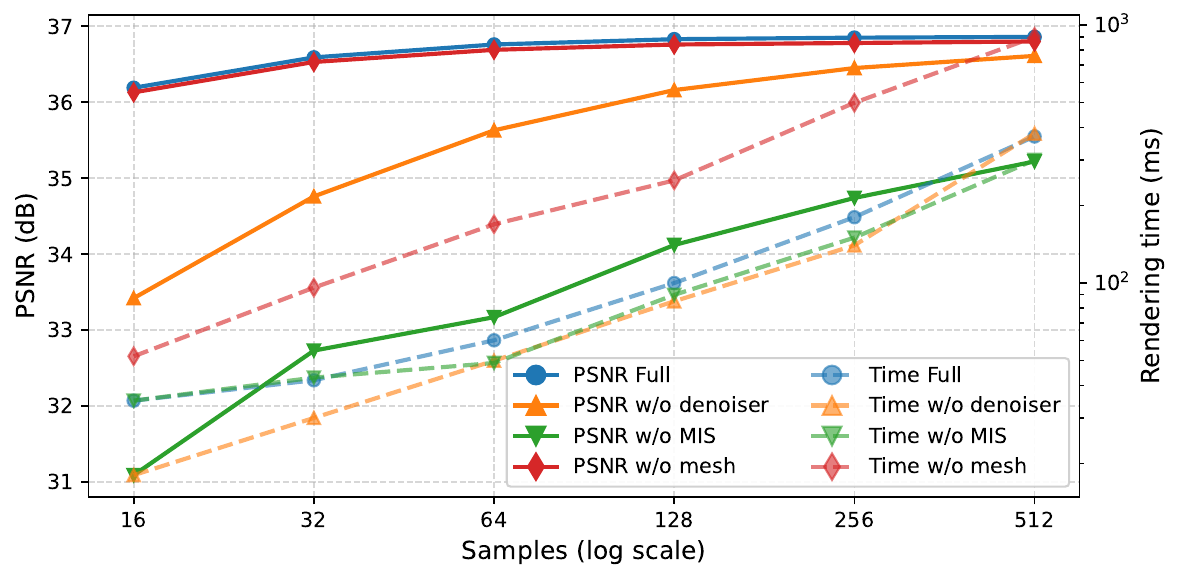} 
    \caption{
    \textbf{Effect of denoiser, MIS, and mesh acceleration on quality and runtime}. It plots PSNR (left axis) and rendering time (right axis, log scale) versus the number of samples for the full method and ablated variants on the Armadillo scene.
    }
    \label{fig:efficiency}
\end{figure}

\noindent\textbf{Efficiency analysis.}
Fig.~\ref{fig:efficiency} shows that the denoiser, MIS, and mesh acceleration substantially reduce rendering time and improve/stabilize PSNR under the same sampling budget, leading to a better quality–efficiency trade-off.


\noindent\textbf{More results.}
To assess the generalization of DR-GS, Fig.~\ref{fig:main_relight_deform} showcases diverse geometric deformations and relighting outcomes under various illumination conditions. These include both bright and dim settings across indoor and outdoor environments, utilizing particle-driven and mesh-driven methods. Further results are available in the supplementary video.

\noindent
\begin{minipage}[c]{0.46\linewidth}
\raggedright
\noindent\textbf{Complex non-manifold geometry.}
Unlike mesh-based methods, GS-based approaches better handle non-manifold topologies. On the TensoIR~\cite{Jin2023TensoIR} (ficus) and Shelly~\cite{adaptiveshells2023} (khady, kitten, pug, wolly) datasets, covering foliage, hair, fur, and porous structures, DR-GS achieves high-fidelity NVS (Fig.~\ref{fig:non-manifold}) where mesh-based methods often fail.
\end{minipage}\hfill
\begin{minipage}[c]{0.52\linewidth}
\centering
\includegraphics[width=\linewidth]{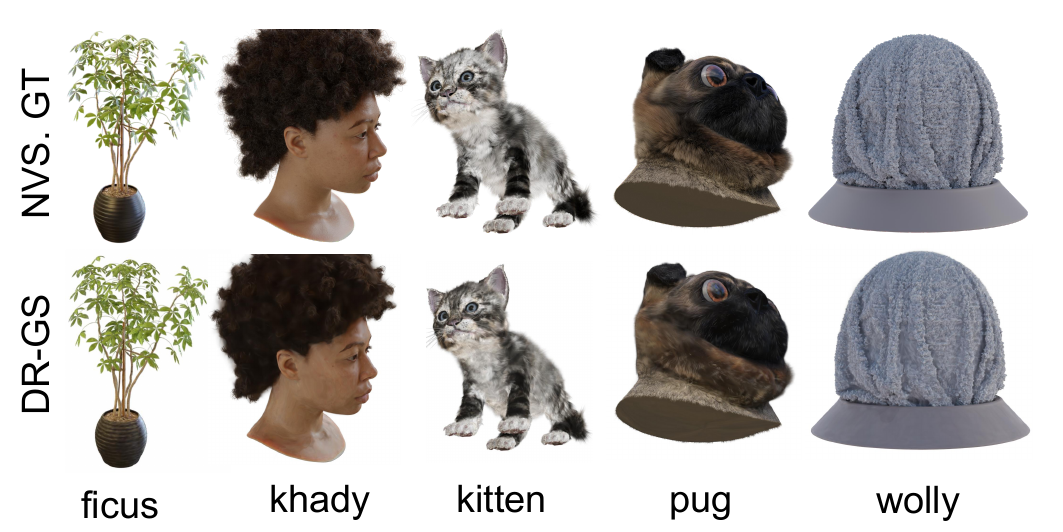}
\captionof{figure}{\textbf{Results of non-manifold geometry}. It demonstrates DR-GS’s ability to deal with intricate structures.}
\label{fig:non-manifold}
\end{minipage}
\section{Conclusion}
This paper presents DR-GS, a unified Gaussian framework for physically-based inverse rendering, relighting, and deformation-aware manipulation. By decoupling geometry, illumination, and materials, DR-GS addresses the unrealistic appearance of snapshots-based methods under geometric and lighting variations, while enabling flexible parameter control. We incorporate MIS and low-sample Monte Carlo estimation to achieve efficient dynamic rendering with high physical fidelity, particularly on glossy surfaces. Supporting both particle- and mesh-based deformation, DR-GS facilitates physical simulation and 3D animation, offering a new pathway for virtual content creation.

\noindent\textbf{Limitation.} While DR-GS eliminates baked-illumination artifacts in dynamic scenes, it introduces a trade-off: solving the rendering equation inherently reduces both training and rendering efficiency compared to vanilla 3DGS.



\section*{Acknowledgements}

This work was supported in part by New Generation Artificial Intelligence-National Science and Technology Major Project (2025ZD0123004), \\Ningbo grant (2025Z038) and National Natural Science Foundation of China (Grant No.62376060).


\bibliographystyle{splncs04}
\bibliography{main}

@String(CVPR  = {IEEE Conf. Comput. Vis. Pattern Recog.})

@String(ICCV  = {Int. Conf. Comput. Vis.})

@String(ECCV  = {Eur. Conf. Comput. Vis.})

@String(NeurIPS = {Adv. Neural Inform. Process. Syst.})

@String(ICLR  = {Int. Conf. Learn. Represent.})

@String(AAAI  = {AAAI})

@String(TOG   = {ACM Trans. Graph.})

@String(TVCG  = {IEEE Trans. Vis. Comput. Graph.})

@String(CVPR  = {CVPR})

@String(ICCV  = {ICCV})

@String(ECCV  = {ECCV})

@String(NeurIPS = {NeurIPS})

@String(ICLR  = {ICLR})

@String(TOG   = {ACM TOG})

@String(TVCG  = {IEEE TVCG})

@Article{3DGS,
      author       = {Kerbl, Bernhard and Kopanas, Georgios and Leimk{\"u}hler, Thomas and Drettakis, George},
      title        = {3D Gaussian Splatting for Real-Time Radiance Field Rendering},
      journal      = {ACM TOG},
      year         = {2023},
}

@inproceedings{2DGS,
    title={2D Gaussian Splatting for Geometrically Accurate Radiance Fields},
    author={Huang, Binbin and Yu, Zehao and Chen, Anpei and Geiger, Andreas and Gao, Shenghua},
    booktitle = {SIGGRAPH},
    year      = {2024},
}

@inproceedings{mani-gs,
  title={Mani-GS: Gaussian Splatting Manipulation with Triangular Mesh},
  author={Gao, Xiangjun and Li, Xiaoyu and Zhuang, Yiyu and Zhang, Qi and Hu, Wenbo and Zhang, Chaopeng and Yao, Yao and Shan, Ying and Quan, Long},
  booktitle={CVPR},
  year={2025}
}

@article {GaussianMesh,
    author = {Gao, Lin and Yang, Jie and Zhang, Botao and Sun, Jiamu and Yuan, Yujie and Fu, Hongbo and Lai, Yu-Kun},
    title = {Real-time Large-scale Deformation of Gaussian Splatting},
    journal = {ACM TOG},
    year = {2024},
}

@article{nero,
  title={Nero: Neural geometry and brdf reconstruction of reflective objects from multiview images},
  author={Liu, Yuan and Wang, Peng and Lin, Cheng and Long, Xiaoxiao and Wang, Jiepeng and Liu, Lingjie and Komura, Taku and Wang, Wenping},
  journal={ACM Transactions on Graphics (ToG)},
  year={2023},
  publisher={ACM New York, NY, USA}
}

@inproceedings{ref-gaussian,
  title={Reflective Gaussian Splatting},
  author={Yao, Yuxuan and Zeng, Zixuan and Gu, Chun and Zhu, Xiatian and Zhang, Li},
  booktitle={ICLR},
  year={2025},
}

@inproceedings{PhysGaussian,
      title={PhysGaussian: Physics-Integrated 3D Gaussians for Generative Dynamics}, 
      author={Xie, Tianyi and Zong, Zeshun and Qiu, Yuxing and Li, Xuan and Feng, Yutao and Yang, Yin and Jiang, Chenfanfu},
      booktitle={CVPR},
      year={2024},
}

@inproceedings{sugar,
    title={SuGaR: Surface-Aligned Gaussian Splatting for Efficient 3D Mesh Reconstruction and High-Quality Mesh Rendering},
    author={Gu{\'e}don, Antoine and Lepetit, Vincent},
    booktitle={CVPR},
    year={2024}
    }

@INPROCEEDINGS{gsp,
  title={Gaussian Splashing: Unified Particles for Versatile Motion Synthesis and Rendering},
  author={Feng, Yutao and Feng, Xiang and Shang, Yintong and Jiang, Ying and Yu, Chang and Zong, Zeshun and Shao, Tianjia and Wu, Hongzhi and Zhou, Kun and Jiang, Chenfanfu and Yang, Yin},
  booktitle={CVPR},
  year={2025}
}

@ARTICLE{ssim,
  author={Zhou Wang and Bovik, A.C. and Sheikh, H.R. and Simoncelli, E.P.},
  journal={IEEE Transactions on Image Processing}, 
  title={Image quality assessment: from error visibility to structural similarity}, 
  year={2004}
}

@INPROCEEDINGS{LPIPS,
  author={Zhang, Richard and Isola, Phillip and Efros, Alexei A. and Shechtman, Eli and Wang, Oliver},
  booktitle={CVPR}, 
  title={The Unreasonable Effectiveness of Deep Features as a Perceptual Metric}, 
  year={2018}
}

@article{GMLS,
  title={Unified simulation of elastic rods, shells, and solids},
  author={Martin, Sebastian and Kaufmann, Peter and Botsch, Mario and Grinspun, Eitan and Gross, Markus},
  journal={ACM Transactions on Graphics (TOG)},
  year={2010},
  publisher={ACM New York, NY, USA}
}

@article{3dgrt,
  title={3d gaussian ray tracing: Fast tracing of particle scenes},
  author={Moenne-Loccoz, Nicolas and Mirzaei, Ashkan and Perel, Or and De Lutio, Riccardo and Martinez Esturo, Janick and State, Gavriel and Fidler, Sanja and Sharp, Nicholas and Gojcic, Zan},
  journal={ACM Transactions on Graphics (TOG)},
  volume={43},
  number={6},
  pages={1--19},
  year={2024},
  publisher={ACM New York, NY, USA}
}

@article{optix,
author = {Parker, Steven G. and Bigler, James and Dietrich, Andreas and Friedrich, Heiko and Hoberock, Jared and Luebke, David and McAllister, David and McGuire, Morgan and Morley, Keith and Robison, Austin and Stich, Martin},
title = {OptiX: a general purpose ray tracing engine},
year = {2010},
journal = {ACM TOG},
}

@inproceedings{IRGS,
  title={IRGS: Inter-Reflective Gaussian Splatting with 2D Gaussian Ray Tracing},
  author={Gu, Chun and Wei, Xiaofei and Zeng, Zixuan and Yao, Yuxuan and Zhang, Li},
  booktitle={CVPR},
  year={2025},
}

@inproceedings{mis,
author = {Veach, Eric and Guibas, Leonidas J.},
title = {Optimally combining sampling techniques for Monte Carlo rendering},
year = {1995},
booktitle = {Proceedings of the 22nd Annual Conference on Computer Graphics and Interactive Techniques}
}

@inproceedings{gs-ror2,
    title={GS-ROR$^2$: Bidirectional-guided 3DGS and SDF for Reflective Object Relighting and Reconstruction}, 
    author={Zuo-Liang Zhu and Beibei Wang and Jian Yang},
    year={2025},
    booktitle={ACM TOG}
}

@inproceedings{nerv,
  title={NeRV: Neural Reflectance and Visibility Fields for Relighting and View Synthesis},
  author={Pratul P. Srinivasan and Boyang Deng and Xiuming Zhang and Matthew Tancik and Ben Mildenhall and Jonathan T. Barron},
  booktitle={CVPR},
  year={2021}}

@inproceedings{nerd,
  title         = {NeRD: Neural Reflectance Decomposition from Image Collections},
  author        = {Boss, Mark and Braun, Raphael and Jampani, Varun and Barron, Jonathan T. and Liu, Ce and Lensch, Hendrik P.A.},
  booktitle     = {ICCV},
  year          = {2021},
}

@inproceedings{NeILF++,
  title={Neilf++: Inter-reflectable light fields for geometry and material estimation},
  author={Zhang, Jingyang and Yao, Yao and Li, Shiwei and Liu, Jingbo and Fang, Tian and McKinnon, David and Tsin, Yanghai and Quan, Long},
  booktitle={Proceedings of the IEEE/CVF International Conference on Computer Vision},
  pages={3601--3610},
  year={2023}
}

@inproceedings{SAMURAI,
      title={SAMURAI: Shape And Material from Unconstrained Real-world Arbitrary Image collections}, 
      author={Mark Boss and Andreas Engelhardt and Abhishek Kar and Yuanzhen Li and Deqing Sun and Jonathan T. Barron and Hendrik P. A. Lensch and Varun Jampani},
      year={2022},
      booktitle     = {NeurIPS}
    
}

@inproceedings{Gs-ir,
  title={Gs-ir: 3d gaussian splatting for inverse rendering},
  author={Liang, Zhihao and Zhang, Qi and Feng, Ying and Shan, Ying and Jia, Kui},
  booktitle={CVPR},
  year={2024}
}

@inproceedings{R3DG,
    author    = {Gao, Jian and Gu, Chun and Lin, Youtian and Zhu, Hao and Cao, Xun and Zhang, Li and Yao, Yao},
    title     = {Relightable 3D Gaussian: Real-time Point Cloud Relighting with BRDF Decomposition and Ray Tracing},
    booktitle   = {ECCV},
    year      = {2024},
}

@ARTICLE{GIR,
  author={Shi, Yahao and Wu, Yanmin and Wu, Chenming and Liu, Xing and Zhao, Chen and Feng, Haocheng and Zhang, Jian and Zhou, Bin and Ding, Errui and Wang, Jingdong},
  journal={T-PAMI}, 
  title={GIR: 3D Gaussian Inverse Rendering for Relightable Scene Factorization}, 
  year={2025},

}

@inproceedings{SVG-IR,
    author={Hanxiao Sun and Yupeng Gao and Jin Xie and Jian Yang and Beibei Wang},
    title={SVG-IR: Spatially-varying Gaussian Splatting for Inverse Rendering},
    year={2025},
    booktitle={CVPR},
    }

@inproceedings{GI-GS,
      title={GI-GS: Global Illumination Decomposition on Gaussian Splatting for Inverse Rendering}, 
      author={Hongze Chen and Zehong Lin and Jun Zhang},
      booktitle={ICLR},
      year={2025},
}

@inproceedings{tensoflow,
  title={TensoFlow: Tensorial Flow-based Sampler for Inverse Rendering},
  author={Gu, Chun and Wei, Xiaofei and Zhang, Li and Zhu, Xiatian},
  booktitle={CVPR},
  year={2025},
}

@article{zhu2024multitimesmontecarlorendering,
  title={Multi-times monte carlo rendering for inter-reflection reconstruction},
  author={Zhu, Tengjie and Chen, Zhuo and Gao, Jingnan and Yan, Yichao and Yang, Xiaokang},
  journal={NeurIPS},
  year={2024}
}

@inproceedings{4DGS,
    author    = {Wu, Guanjun and Yi, Taoran and Fang, Jiemin and Xie, Lingxi and Zhang, Xiaopeng and Wei, Wei and Liu, Wenyu and Tian, Qi and Wang, Xinggang},
    title     = {4D Gaussian Splatting for Real-Time Dynamic Scene Rendering},
    booktitle = {CVPR},
    year      = {2024},
    }

@InProceedings{NeuMA,
    author    = {Cao, Junyi and Guan, Shanyan and Ge, Yanhao and Li, Wei and Yang, Xiaokang and Ma, Chao},
    title     = {Neu{MA}: Neural Material Adaptor for Visual Grounding of Intrinsic Dynamics},
    booktitle = {NeurIPS},
    year      = {2024}
}

@inproceedings{physdreamer,
    title={{PhysDreamer}: Physics-Based Interaction with 3D Objects via Video Generation},
    author={Tianyuan Zhang and Hong-Xing Yu and Rundi Wu and Brandon Y. Feng and Changxi Zheng and Noah Snavely and Jiajun Wu and William T. Freeman},
    booktitle={ECCV},
    year={2024},
    }

@inproceedings{dreamphysics,
  title={DreamPhysics: Learning Physical Properties of Dynamic 3D Gaussians with Video Diffusion Priors},
  author={Huang, Tianyu and Zeng, Yihan and Li, Hui and Zuo, Wangmeng and Lau, Rynson WH},
  booktitle={AAAI},
  year={2025}
}

@article{physics3d,
  title={Physics3D: Learning Physical Properties of 3D Gaussians via Video Diffusion},
  author={Liu, Fangfu and Wang, Hanyang and Yao, Shunyu and Zhang, Shengjun and Zhou, Jie and Duan, Yueqi},
  journal={arXiv preprint arXiv:2406.04338},
  year={2024}
}

@article{efficient,
  title={Efficient Physics Simulation for 3D Scenes via MLLM-Guided Gaussian Splatting},
  author={Zhao, Haoyu and Wang, Hao and Zhao, Xingyue and Fei, Hao and Wang, Hongqiu and Long, Chengjiang and Zou, Hua},
  journal={arXiv preprint arXiv:2411.12789},
  year={2024}
}

@inproceedings{featuresplatting,
      title={Language-Driven Physics-Based Scene Synthesis and Editing via Feature Splatting},
      author={Ri-Zhao Qiu and Ge Yang and Weijia Zeng and Xiaolong Wang},
      booktitle={ECCV},
      year={2024}
    }

@article{games,
  title={Games: Mesh-based adapting and modification of gaussian splatting},
  author={Waczy{\'n}ska, Joanna and Borycki, Piotr and Tadeja, S{\l}awomir and Tabor, Jacek and Spurek, Przemys{\l}aw},
  journal={arXiv preprint arXiv:2402.01459},
  year={2024}
}

@inproceedings{vr-gs,
      title={VR-GS: A Physical Dynamics-Aware Interactive Gaussian Splatting System in Virtual Reality},
      author={Jiang, Ying and Yu, Chang and Xie, Tianyi and Li, Xuan and Feng, Yutao and Wang, Huamin and Li, Minchen and Lau, Henry and Gao, Feng and Yang, Yin and Jiang, Chenfanfu},
      booktitle={SIGGRAPH},
      year={2024},
}

@article{vrdoh,
author = {Zhaofeng Luo and Zhitong Cui and Shijian Luo and Mengyu Chu and Minchen Li},
title = {VR-Doh: Hands-on 3D Modeling in Virtual Reality},
year = {2025},
journal = {ACM TOG},
}

@InProceedings{deformable3dgs,
    title={Deformable 3D Gaussians for High-Fidelity Monocular Dynamic Scene Reconstruction},
    author={Yang, Ziyi and Gao, Xinyu and Zhou, Wen and Jiao, Shaohui and Zhang, Yuqing and Jin, Xiaogang},
    booktitle={CVPR},
    year={2024}
}

@InProceedings{SpacetimeGS,
    author    = {Li, Zhan and Chen, Zhang and Li, Zhong and Xu, Yi},
    title     = {Spacetime Gaussian Feature Splatting for Real-Time Dynamic View Synthesis},
    booktitle = {CVPR},
    year      = {2024},
}

@inproceedings{gaussianavatars,
  title={Gaussianavatars: Photorealistic head avatars with rigged 3d gaussians},
  author={Qian, Shenhan and Kirschstein, Tobias and Schoneveld, Liam and Davoli, Davide and Giebenhain, Simon and Nie{\ss}ner, Matthias},
  booktitle={CVPR},
  year={2024}
}

@inproceedings{gaussianavatar,
        title={GaussianAvatar: Towards Realistic Human Avatar Modeling from a Single Video via Animatable 3D Gaussians},
        author={Hu, Liangxiao and Zhang, Hongwen and Zhang, Yuxiang and Zhou, Boyao and Liu, Boning and Zhang, Shengping and Nie, Liqiang},
        booktitle={CVPR},
        year={2024}
}

@article{anigaussiananimatablegaussianavatar,
  title={AniGaussian: Animatable Gaussian Avatar with Pose-guided Deformation},
  author={Mengtian Li and Shengxiang Yao and Chen Kai and Zhifeng Xie and Keyu Chen and Yu-Gang Jiang},
  journal={ArXiv},
  year={2025},
}

@inproceedings{BezierGS,
  title={BézierGS: Dynamic Urban Scene Reconstruction with Bézier Curve Gaussian Splatting},
  author={Ma, Zipei and Jiang, Junzhe and Chen, Yurui and Zhang, Li},
  booktitle={ICCV},
  year={2025},
}

@article{pvg,
  title={Periodic Vibration Gaussian: Dynamic Urban Scene Reconstruction and Real-time Rendering},
  author={Chen, Yurui and Gu, Chun and Jiang, Junzhe and Zhu, Xiatian and Zhang, Li},
  journal={arXiv:2311.18561},
  year={2023},
}

@inproceedings{yang2023gs4d,
  title={Real-time Photorealistic Dynamic Scene Representation and Rendering with 4D Gaussian Splatting},
  author={Yang, Zeyu and Yang, Hongye and Pan, Zijie and Zhang, Li},
  booktitle={ICLR},
  year={2024}
}

@article{glossygs2024,
      author={Lai, Shuichang and Huang, Letian and Guo, Jie and Cheng, Kai and Pan, Bowen and Long, Xiaoxiao and Lyu, Jiangjing and Lv, Chengfei and Guo, Yanwen},
      journal={TVCG}, 
      title={GlossyGS: Inverse Rendering of Glossy Objects With 3D Gaussian Splatting}, 
      year={2025},
}

@inproceedings{PRTGS2024,
author = {Guo, Yijia and Bai, Yuanxi and Hu, Liwen and Guo, Ziyi and Liu, Mianzhi and Cai, Yu and Huang, Tiejun and Ma, Lei},
title = {PRTGS: Precomputed Radiance Transfer of Gaussian Splats for Real-Time High-Quality Relighting},
year = {2024},
booktitle = {MM} ,
}

@article{2020nerf,
  title={Nerf: Representing scenes as neural radiance fields for view synthesis},
  author={Mildenhall, Ben and Srinivasan, Pratul P and Tancik, Matthew and Barron, Jonathan T and Ramamoorthi, Ravi and Ng, Ren},
  journal={Communications of the ACM},
  year={2021},
  publisher={ACM New York, NY, USA}
}

@inproceedings{lu2024manigaussian,
      title={ManiGaussian: Dynamic Gaussian Splatting for Multi-task Robotic Manipulation}, 
      author={Lu, Guanxing and Zhang, Shiyi and Wang, Ziwei and Liu, Changliu and Lu, Jiwen and Tang, Yansong},
      booktitle={ECCV},
      year={2024}
}

@article{yu2025real2render2realscalingrobotdata,
  title={Real2render2real: Scaling robot data without dynamics simulation or robot hardware},
  author={Yu, Justin and Fu, Letian and Huang, Huang and El-Refai, Karim and Ambrus, Rares Andrei and Cheng, Richard and Irshad, Muhammad Zubair and Goldberg, Ken},
  journal={arXiv preprint arXiv:2505.09601},
  year={2025}
}

@inproceedings{shorinwa2024splatmovermultistageopenvocabularyrobotic,
      title={Splat-MOVER: Multi-Stage, Open-Vocabulary Robotic Manipulation via Editable Gaussian Splatting},
      author={Shorinwa, Ola and Tucker, Johnathan and Smith, Aliyah and Swann, Aiden and Chen, Timothy and Firoozi, Roya and Kennedy, Monroe David and Schwager, Mac},
      booktitle={8th Annual Conference on Robot Learning},
      year={2024}
    }

@inproceedings{ji2024graspsplats,
    title={GraspSplats: Efficient Manipulation with 3D Feature Splatting}, 
    author={Mazeyu Ji and Ri-Zhao Qiu and Xueyan Zou and Xiaolong Wang},
    booktitle={CoRL},
    year={2024}
}

@article{zheng2024gaussiangrasper,
  title={Gaussiangrasper: 3d language gaussian splatting for open-vocabulary robotic grasping},
  author={Zheng, Yuhang and Chen, Xiangyu and Zheng, Yupeng and Gu, Songen and Yang, Runyi and Jin, Bu and Li, Pengfei and Zhong, Chengliang and Wang, Zengmao and Liu, Lina and others},
  journal={IEEE Robotics and Automation Letters},
  year={2024},
  publisher={IEEE}
}

@InProceedings{guo2025pgc,
  title={PGC: Physics-Based Gaussian Cloth from a Single Pose},
  author={Guo, Michelle and Chiang, Matt Jen-Yuan and Santesteban, Igor and Sarafianos, Nikolaos and Chen, Hsiao-yu and Halimi, Oshri and Bo{\v{z}}i{\v{c}}, Alja{\v{z}} and Saito, Shunsuke and Wu, Jiajun and Liu, C Karen and Stuyck, Tuur and Larionov, Egor},
  booktitle={CVPR},
  year={2025}
}

@InProceedings{li2025wonderplay,
    title     = {WonderPlay: Dynamic 3D Scene Generation from a Single Image and Actions},
    author    = {Li, Zizhang and Yu, Hong-Xing and Liu, Wei and Yang, Yin and Herrmann, Charles and Wetzstein, Gordon and Wu, Jiajun},
    booktitle = {ICCV},
    year      = {2025},
}

@article{Heitz2018GGX,
  author =       {Eric Heitz}, 
  title =        {Sampling the GGX Distribution of Visible Normals},
  year =         2018,
  journal =      {JCGT},
}

@book{PBR,
author = {Pharr, Matt and Jakob, Wenzel and Humphreys, Greg},
title = {Physically Based Rendering: From Theory to Implementation},
year = {2016},
address = {San Francisco, CA, USA},
edition = {3rd},
}

@inproceedings{rendering_equation,
author = {Kajiya, James T.},
title = {The rendering equation},
year = {1986},
booktitle = {Proceedings of the 13th Annual Conference on Computer Graphics and Interactive Techniques},

}

@article{a_reflectance_model_for_computer_graphics,
author = {Cook, R. L. and Torrance, K. E.},
title = {A Reflectance Model for Computer Graphics},
year = {1982},
journal = {ACM TOG},
}

@inproceedings{SVGF,
author = {Schied, Christoph and Kaplanyan, Anton and Wyman, Chris and Patney, Anjul and Chaitanya, Chakravarty R. Alla and Burgess, John and Liu, Shiqiu and Dachsbacher, Carsten and Lefohn, Aaron and Salvi, Marco},
title = {Spatiotemporal variance-guided filtering: real-time reconstruction for path-traced global illumination},
year = {2017},
booktitle = {Proceedings of High Performance Graphics},
}

@article{Flexicubes,
author = {Shen, Tianchang and Munkberg, Jacob and Hasselgren, Jon and Yin, Kangxue and Wang, Zian and Chen, Wenzheng and Gojcic, Zan and Fidler, Sanja and Sharp, Nicholas and Gao, Jun},
title = {Flexible Isosurface Extraction for Gradient-Based Mesh Optimization},
year = {2023},
journal = {ACM TOG},
}

@inproceedings{Jin2023TensoIR,
  title={TensoIR: Tensorial Inverse Rendering},
  author={Jin, Haian and Liu, Isabella and Xu, Peijia and Zhang, Xiaoshuai and Han, Songfang and Bi, Sai and Zhou, Xiaowei and Xu, Zexiang and Su, Hao},
  booktitle={CVPR},
  year={2023}
}

@article{adaptiveshells2023,
  author = {Zian Wang and Tianchang Shen and Merlin Nimier-David and Nicholas Sharp and Jun Gao and Alexander Keller and Sanja Fidler and Thomas M\"uller and Zan Gojcic},
  title = {Adaptive Shells for Efficient Neural Radiance Field Rendering},
  journal = {ACM Trans. Graph.},
  issue_date = {December 2023},
  volume = {42},
  number = {6},
  year = {2023},
  articleno = {259},
  numpages = {15},
  url = {https://doi.org/10.1145/3618390},
  doi = {10.1145/3618390},
  publisher = {ACM},
  address = {New York, NY, USA}
}
\end{document}